\documentclass[10pt,journal,compsoc]{IEEEtran}
\usepackage[nocompress]{cite}

\usepackage{hyperref}
\usepackage[english]{babel}
\usepackage{amsmath}
\usepackage{amssymb}
\usepackage[table]{xcolor}
\usepackage{tikz}
\usepackage{pgfplots}
\usepackage{xifthen}
\usepackage{caption}
\usepackage{subcaption}
\usepackage{xspace}
\usepackage{enumitem}
\usepackage[textwidth=1.7cm,textsize=small,disable]{todonotes}
\usepackage{import}
\usepackage{dcolumn}
\usepackage{array}
\usepackage{textcomp}
\usepackage{newtxmath}

\makeatletter
\newcolumntype{d}{D{.}{.}{-1} }
\newcolumntype{B}[3]{>{\boldmath\DC@{#1}{#2}{#3} }c<{\DC@end} }
\makeatother

\def\jantodo#1{\todo[color=orange!50!white]{#1}\xspace}

\usetikzlibrary{calc}
\usetikzlibrary{positioning}

\usetikzlibrary{arrows}
\usetikzlibrary{backgrounds}
\usetikzlibrary{calc}
\usetikzlibrary{chains}
\usetikzlibrary{fit}
\usetikzlibrary{matrix}
\usetikzlibrary{mindmap}
\usetikzlibrary{pgfplots.colormaps}
\usetikzlibrary{positioning}
\usetikzlibrary{shadows}
\usetikzlibrary{shapes}
\usetikzlibrary{trees}

\tikzstyle{inactive}=
  [draw=gray,text=gray]

\tikzstyle{data}=
  [rectangle,rounded corners,draw,fill=gray!10!white]
\tikzstyle{method}=
  [rounded rectangle,draw]

\tikzstyle{image}=
  [rectangle,draw,fill=white,inner sep=2pt]

\tikzstyle{process}=
  [single arrow,shading=axis,left color=orange!50!white, right color=orange!40!white,draw=orange!50!white!80!black,inner sep=2pt,minimum height=8mm,single arrow head extend=2mm]

\tikzstyle{voxel}=
  [circle,fill=black!90!white,inner sep=2pt]

\tikzstyle{gt_region}=
  [line width=10mm,line cap=round,line join=round]

\tikzstyle{affinity}=
  [line width=1pt,draw=black!90!white]

\tikzstyle{affinity_value}=
  [opacity=0] % don't draw affinity values

\tikzstyle{mst}=
  [line width=11pt,line cap=round,line join=round,draw=gray!75!white]

\tikzstyle{kernel}=
  [draw=gray,fill=blue!10!white]

\tikzstyle{flow}=
  [->,draw=gray,thin]

\tikzstyle{conv_pass}=
  [single arrow,shading=axis,left color=blue!50!white, right color=blue!10!white,draw=blue!50!white!70!black, inner sep=0]

\tikzstyle{copy_pass}=
  [single arrow,shading=axis,left color=purple!50!white, right color=purple!10!white,draw=purple!50!white!70!black, inner sep=0]

\tikzstyle{max_pool_pass}=
  [single arrow,shading=axis,left color=orange!50!white, right color=orange!10!white,draw=orange!50!white!70!black, inner sep=0]

\tikzstyle{upsampling_pass}=
  [single arrow,shading=axis,left color=brown!50!white, right color=brown!10!white,draw=brown!50!white!70!black, inner sep=0]

\tikzstyle{bb}=
  [inner sep=0,outer sep=0]

\tikzstyle{malis_highlight_node}=
  [circle,inner sep=2mm,fill=blue!25!white]

\tikzstyle{malis_highlight_edge}=
  [line width=5mm,draw=blue!25!white,line cap=round,line join=round]

\tikzstyle{unet_l1}=[]
\tikzstyle{unet_l2}=
  [scale=0.7,transform shape]
\tikzstyle{unet_l3}=
  [scale=0.5,transform shape]
\tikzstyle{unet_l4}=
  [scale=0.5,transform shape]

\tikzstyle{unet_annotation}=
  [circle,fill=white,draw=black,minimum width=4mm]

\tikzstyle{frame}=
  [rectangle,draw,fill=white]

\tikzstyle{region}=
  [circle,draw,fill=gray!10!white,minimum width=0.5cm]

\tikzstyle{match}=
  [line width=2pt,blue!50!white]

\tikzstyle{synapse}=
  [->,>=stealth',thick]

\tikzstyle{merge_highlight}=
  [line width=10mm,draw=blue!25!white,line cap=round,line join=round]

\tikzstyle{queue_element}=
  [rectangle,draw=orange!50!white!50!black,fill=orange!50!white]

\tikzstyle{adjacency}=
  [draw=black]
\tikzstyle{subset}=
  [draw=gray,thick,->]

\newcommand{\getzoomfactor}{%
\pgfgettransformentries{\myxscale}{\@tempa}{\@tempa}{\myyscale}{\@tempa}{\@tempa}
\gdef\zoomfactor{\myxscale}
}

\def\scalebar#1#2{%
  \fill[black] (0,0) rectangle ($(#1,-#1*0.2)$);
  \path (0,0) -- node[above] {#2} (#1,0);
}

\pgfplotsset{compat=1.13}
\usetikzlibrary{pgfplots.groupplots}
\usepgfplotslibrary{statistics}

\pgfplotsset{
  errors/.style={
    stack plots=y,
    area style,
    enlarge x limits=false,
    xmajorgrids=true,
    ymajorgrids=true,
    yminorgrids=true,
    legend reversed
  }
}

\pgfplotsset{
  discard if not/.style 2 args={
    x filter/.code={
      \edef\tempa{\thisrow{#1}}
      \edef\tempb{#2}
      \ifx\tempa\tempb
      \else
        
      \fi
    }
  }
}

\def\figref#1{Fig.~\ref{#1}}
\def\subfigref#1{(\subref{#1})}
\def\secref#1{Section~\ref{#1}}
\def\tabref#1{Table~\ref{#1}}
\def\eqref#1{(\ref{#1})}

\def\ie{\emph{i.e.}\xspace}

\def\malis{\textsc{Malis}\xspace}
\def\mala{\textsc{Mala}\xspace}
\def\constmalis{constrained \malis\xspace}
\def\celis{\textsc{Celis}\xspace}
\def\multicut{\textsc{MultiCut}\xspace}
\def\unet{\textsc{U-Net}\xspace}
\def\gala{\textsc{Gala}\xspace}

\def\cremi{\textsc{Cremi}\xspace}
\def\segem{\textsc{SegEm}\xspace}
\def\fib{\textsc{Fib-25}\xspace}

\def\caffe{\textsc{caffe}\xspace}

\DeclareMathOperator{\mm}{mm}
\DeclareMathOperator{\mtp}{mtp}
\def\seg{s}
\def\aff{a}
\def\malisloss{L}
\def\affloss{l}
\def\weight{w}

\newcommand{\argmin}[1]{\mathop{\arg\min}_{#1}\hspace{0.5em}}

\newcommand{\argmax}[1]{\mathop{\arg\max}_{#1}\hspace{0.5em}}

\newcommand{\x}{\ensuremath{\text{\texttimes}}}

\begin{document}

%\title{A Deep Structured Learning Approach Towards Automating Connectome Reconstruction from 3D Electron Micrographs}%
%\title{A Structured-Loss Deep Learning Approach Towards Automating Connectome Reconstruction from 3D Electron Micrographs}%
%\title{A Structured Loss for Affinity Prediction to Automate Connectome Reconstruction from 3D Electron Micrographs}%
%\title{A Structured Loss for Affinity Prediction Towards Automating Connectome Reconstruction from 3D Electron Micrographs}%
%\title{\rererevised A Structured Loss for Neuron Segmentation Towards Automating Connectome Reconstruction from 3D Electron Micrographs}%
\title{\rererevised Large Scale Image Segmentation with Structured Loss based Deep Learning for Connectome Reconstruction}%

\author{
  Jan Funke\thanks{$^\star$ these authors contributed equally}$^{\star}$,
  Fabian David Tschopp$^{\star}$,
  William Grisaitis,
  Arlo Sheridan,\\
  Chandan Singh,
  Stephan Saalfeld,
  Srinivas C. Turaga
}

\def\revised{} % uncomment for final version
\def\rerevised{} % uncomment for final version
\def\rererevised{} % uncomment for final version

%\thanks{$^{\star}$ authors contributed equally}
%\author{
  %Jan Funke$^{\star}$\\
  %Institut de Rob\`otica i Inform\`atica Industrial\\
  %UPC/CSIC Barcelona\\
  %\texttt{jfunke@iri.upc.edu} \\
  %%\And
  %Fabian David Tschopp$^{\star}$\\
  %Institute of Neuroinformatics\\
  %University of Zurich and ETH Zurich\\
  %\texttt{tschopfa@student.ethz.ch} \\
  %%\And
  %William Grisaitis\\
  %Janelia Research Campus \\
  %Howard Hughes Medical Instutute \\
  %\texttt{grisaitisw@janelia.hhmi.org} \\
  %%\And
  %Chandan Singh\\
  %Janelia Research Campus \\
  %Howard Hughes Medical Instutute \\
  %\texttt{chandan\_singh@berkeley.edu} \\
  %%\And
  %Stephan Saalfeld \\
  %Janelia Research Campus \\
  %Howard Hughes Medical Instutute \\
  %\texttt{saalfelds@janelia.hhmi.org} \\
  %%\And
  %Srinivas C. Turaga \\
  %Janelia Research Campus \\
  %Howard Hughes Medical Instutute \\
  %\texttt{turagas@janelia.hhmi.org} \\
%}

\maketitle

\begin{abstract}

%Three dimensional electron microscopy can be used to image brain tissue with unprecedented resolution and scale, enabling the reconstruction of neural circuitry at the synaptic scale. We present a deep learning system for neuron segmentation which improves significantly upon state of the art, towards the automation of tracing neurons in electron micrographs.

  % What do we present?
  %
  We present a {\rererevised method combining affinity prediction with region
  agglomeration, which} improves significantly upon the state of the art of
  neuron segmentation from electron microscopy (EM) in accuracy and
  scalability.
  % words: 31
  %
  % What are the parts?
  %
  Our method consists of a 3D \unet, trained to predict affinities between
  voxels, followed by iterative region agglomeration. We train using a
  {\rererevised structured loss based on \malis}, encouraging topologically
  correct segmentations {\revised obtained from affinity thresholding}.
  % words: 35
  %
  Our extension consists of two parts: {\rerevised First, we present a
  quasi-linear method to compute the loss gradient, improving over the original
  quadratic algorithm.}
  % words: 23
  %
  Second, we compute the gradient in two separate passes to avoid spurious
  gradient contributions in early training stages.
  % words: 18
  %
  Our predictions are accurate enough that simple learning-free
  percentile-based agglomeration outperforms more involved methods used earlier
  on inferior predictions.
  % words: 19
  %
  % What are the results?
  %
  We present results on three diverse EM datasets, achieving relative
  improvements over previous results of 27\%, 15\%, and 250\%.
  % words: 19
  %
  Our findings suggest that a single method can be applied to both nearly
  isotropic block-face EM data and anisotropic serial sectioned EM data.
  % words: 23
  %
  The runtime of our method scales linearly with the size of the volume and
  achieves a throughput of $\sim2.6$ seconds per megavoxel, qualifying our
  method for the processing of very large datasets.
  % words: 33
  %
  % total: 201 (somehow counted as 199 by the submission form)

\end{abstract}

\section{Introduction}

\begin{figure*}[t]
  \begin{subfigure}[t]{\textwidth}
    \centerline{\subimport{../figures/}{pipeline_overview}}
    \vspace{-5mm}
    %\caption{Overview.}
    %\label{fig:method:overview:pipeline}
    \vspace{2mm}
  \end{subfigure}
  \begin{subfigure}{0.5\textwidth}
    \centerline{\hspace{13mm}\includegraphics[width=1.0\textwidth]{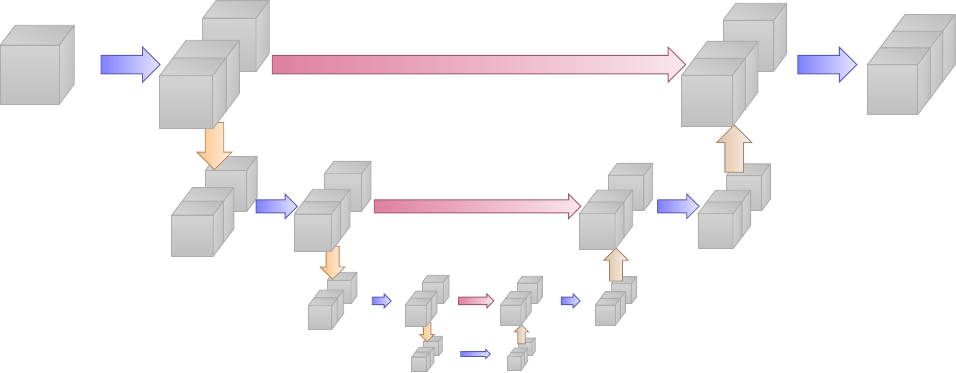}}
    \caption{3D \unet.}
    \label{fig:method:overview:unet}
  \end{subfigure}
  \\
  \begin{subfigure}{0.5\textwidth}
    \vspace{3mm}
    \centerline{\scalebox{0.9}{\subimport{../figures/}{affinity_anisotropy_example}}}
    \vspace{-2mm}
    \caption{Inter-voxel affinities.}
    \label{fig:method:overview:affinities}
  \end{subfigure}
  \begin{subfigure}{0.5\textwidth}
    \vspace{-17mm}
    \centerline{\hspace{2mm}\scalebox{0.95}{\begin{tikzpicture}

  %%%%%%%%%
  % NODES %
  %%%%%%%%%

  \node[region] (A) at (0,0) {A};
  \node[region] (B) at (1,0) {B};
  \node[region] (C) at (2,0) {C};
  \node[region] (D) at (3,0) {D};

  \node[region] (C') at (2,1.5) {C};
  \node[region] (D') at (3,1.5) {D};
  \node[region] (A') at (0.5,1.5) {A};

  \node[region] (C'') at (2.5,3) {C};
  \node[region] (A'') at (0.5,3) {A};

  %%%%%%%%%%%%%%%
  % ADJACENCIES %
  %%%%%%%%%%%%%%%

  \draw[adjacency] (A) -- node[above] {a} (B);
  \draw[adjacency] (B) -- node[above] {b} (C);
  \draw[adjacency] (C) -- node[above] {c} (D);
  \draw[adjacency] (A) to [out=320,in=220] node[below] {d} (C);
  \draw[adjacency] (B) to [out=320,in=220] node[below] {e} (D);

  \draw[adjacency] (C') -- node[above] {c} (D');
  \draw[adjacency] (A') -- node[above] {b} (C');
  \draw[adjacency] (A') to [out=320,in=220] node[below] {e} (D');

  \draw[adjacency] (A'') -- node[above] {b} (C'');

  %%%%%%%%%%%
  % SUBSETS %
  %%%%%%%%%%%

  \draw[subset] (A) -- (A');
  \draw[subset] (B) -- (A');
  \draw[subset] (C) -- (C');
  \draw[subset] (D) -- (D');

  \draw[subset] (C') -- (C'');
  \draw[subset] (D') -- (C'');
  \draw[subset] (A') -- (A'');

  %%%%%%%%%
  % QUEUE %
  %%%%%%%%%

  \node (q1) at (4.5,0)
    {[\underline{a},d,c,b,e]};
  \node (q2) at (4.5,1.5)
    {[\textcolor{red}{d},\underline{c},\textcolor{green!50!black}{b},e]};
  \node (q3) at (4.5,3)
    {[\textcolor{green!50!black}{\underline{b}},\textcolor{red}{e}]};

  \draw[->] (q1) -- node[anchor=west] {b,d $\rightarrow$ b} (q2);
  \draw[->] (q2) -- node[anchor=west] {b,e $\rightarrow$ b} (q3);

  %%%%%%%%%%
  % IMAGES %
  %%%%%%%%%%

  \node[image,inner sep=0] (i1) at (-2,0) {
    \begin{tikzpicture}[scale=1.1]
      \clip (0,0) rectangle (1.2,1.2);
      \draw (0.5,-0.5) to [out=0,in=220] node[pos=0.5] (b) {} node[pos=0.6] (a) {} (0.5,1.5);
      \draw (a.center) to [out=45,in=200] (1.5,0.5);
      \draw (-0.5,0.45) to [out=45,in=180] (b.center);
      \node at (0.2,0.2) {A};
      \node at (0.2,0.8) {B};
      \node at (0.8,0.2) {C};
      \node at (0.8,0.85) {D};
    \end{tikzpicture}
  };
  \node[image,inner sep=0] (i2) at (-2,1.5) {
    \begin{tikzpicture}[scale=1.1]
      \clip (0,0) rectangle (1.2,1.2);
      \draw (0.5,-0.5) to [out=0,in=220] node[pos=0.5] (b) {} node[pos=0.6] (a) {} (0.5,1.5);
      \draw (a.center) to [out=45,in=200] (1.5,0.5);
      \node at (0.2,0.5) {A};
      \node at (0.8,0.2) {C};
      \node at (0.8,0.85) {D};
    \end{tikzpicture}
  };
  \node[image,inner sep=0] (i3) at (-2,3) {
    \begin{tikzpicture}[scale=1.1]
      \clip (0,0) rectangle (1.2,1.2);
      \draw (0.5,-0.5) to [out=0,in=220] node[pos=0.5] (b) {} node[pos=0.6] (a) {} (0.5,1.5);
      \node at (0.2,0.5) {A};
      \node at (0.8,0.5) {C};
    \end{tikzpicture}
  };

  \begin{pgfonlayer}{background}
    \draw[line width=5pt,gray!25!white] (i1.center) -- (3.5,0);
    \draw[line width=5pt,gray!25!white] (i2.center) -- (3.5,1.5);
    \draw[line width=5pt,gray!25!white] (i3.center) -- (3.5,3);
  \end{pgfonlayer}

  %%%%%%%%%%%%%
  % THRESHOLD %
  %%%%%%%%%%%%%

  \draw[->] (-1,-0.5) -- node[rotate=90,below] {threshold} (-1,3.5);

  %%%%%%%%%%%%%%%
  % ANNOTATIONS %
  %%%%%%%%%%%%%%%

  \node at (-2,-1.25) {segmentation};
  \node at (1.5,-1.25) {RAG and merge hierarchy};
  \node at (4.5,-1.25) {queue};

\end{tikzpicture}}}
    \vspace{2mm}
    \caption{Percentile agglomeration.}
    \label{fig:method:overview:agglomeration}
  \end{subfigure}
  \caption{
    Overview of our method (top row). Using a 3D \unet
    (\subref{fig:method:overview:unet}), trained with the proposed \constmalis
    loss, we directly predict inter-voxel affinities from volumes of raw data.
    Affinities provide advantages especially in the case of low-resolution data
    (\subref{fig:method:overview:affinities}). In the example shown here, the
    \emph{voxels} cannot be labeled correctly as foreground/background: If A
    were labeled as foreground, it would necessarily merge with the
    regions in the previous and next section. If it were labeled as background,
    it would introduce a split. The labeling of affinities on \emph{edges}
    allows B and C to separate A from adjacent sections, while maintaining
    connectivity inside the region.
    From the predicted affinities, we obtain an over-segmentation that is
    then merged into the final segmentation using a percentile-based
    agglomeration algorithm (\subref{fig:method:overview:agglomeration}).
  }
  \label{fig:method:overview}
\end{figure*}

% What is the context of this paper?
%
Precise reconstruction of neural connectivity is of great importance to
understand the function of biological nervous systems.
  3D electron microscopy (EM) is the only available imaging method with the
  resolution necessary to visualize and reconstruct dense neural morphology
  without ambiguity. At this resolution, however, even moderately small neural
  circuits yield image volumes that are too large for manual reconstruction.
  Therefore, automated methods for neuron tracing are needed to aid human
  analysis.

% What is this paper about?
%
{\rererevised We present a method combining a structured loss for deep learning
based instance separation with subsequent region agglomeration for neuron
segmentation in 3D electron microscopy, which improves significantly upon state
of the art in terms of accuracy and scalability.} For an overview, see
\figref{fig:method:overview}, top row.
%
  % Main contributions
  %   • simple, scalable pipeline
  %   ✔ 3D U-Net together with
  %   ✔ training on MALIS loss
  %     ✔ improved maximin edge scoring
  %   ✔ efficient agglomeration
  %   … improved state-of-the-art on
  %     • CREMI
  %     • SegEM
  %     • FIB-25
  The main components of our method are:
  (1) Prediction of 3D affinity graphs using a 3D \unet
  architecture~\cite{Cicek2016},
  (2) a structured loss based on \malis~\cite{Turaga2009} to train the \unet to
  minimize topological errors,
  and (3) an efficient $O(n)$ agglomeration scheme based on quantiles of
  predicted affinities.

% Detail of contribution "3D U-Net"
%
The choice of using a 3D~\unet architecture to predict voxel affinities is
motivated by two considerations:
  First, {\unet}s have already shown superior performance on the segmentation
  of 2D~\cite{Ronneberger2015} and 3D~\cite{Cicek2016} biomedical image data.
  One of their favourable properties is the multi-scale architecture which
  enables computational and statistical efficiency. Second, {\unet}s
  efficiently predict large regions. This is of particular interest in
  combination with training on the \malis structured loss, for which we need
  affinity predictions in a region.

% Detail of contribution "constrained MALIS"
%
We train our 3D \unet to predict affinities using an extension of the \malis
loss function~\cite{Turaga2009}.
  Like the original \malis loss, we minimize a topological error on hypothetical
  thresholding and connected component analysis on the predicted affinities. We
  extended the original formulation to derive the gradient with respect to all
  predicted affinities (as opposed to sparsely sampling them), leading to
  denser and faster gradient computation.
  Furthermore, we compute the \malis loss in two passes: In the \emph{positive
  pass}, we constrain all predicted affinities between and outside of
  ground-truth regions to be $0$, and in the \emph{negative pass}, we constrain
  affinities inside regions to be $1$ which avoids spurious gradients in early
  training stages.

% Detail of contribution "efficient agglomeration"
%
Although the training is performed assuming subsequent thresholding, we found
iterative agglomeration of \emph{fragments} (or ``supervoxels'') to be more
robust to small errors in the affinity predictions.
  To this end, we extract fragments running a watershed algorithm on the
  predicted affinities. The fragments are then represented in a region
  adjacency graph (RAG), where edges are scored to reflect the predicted
  affinities between adjacent fragments: edges with small scores will be merged
  before edges with high scores.
  %
  %Using a priority queue, we iteratively merge edges with the lowest score,
  %\ie, we merge the two incident regions $u$ and $v$ into one region $u'$.
  %Likewise, edges to common neighbors of $u$ and $v$ are merged.
  %%
  %The new scores of merged edges are computed in a lazy fashion, which has two
  %speed implications: First, the priority queue does not need to be resorted,
  %and second, not every edge needs to be scored.
  %%
  %These optimizations can be made use of if we assume that the new edge score
  %after a merge is larger or equal to the smaller score of the merged edges,
  %which is true for a wide range of scoring functions. Instead of recomputing
  %the score of an edge right after a merge, we mark the edge score as
  %\emph{stale}. Only when we pop a stale edge from the queue we recompute its
  %score and insert it into the queue again. This way we avoid recomputation and
  %update of the queue for edges that do not contribute to the merge
  %hierarchy.
  %
  We discretize edge scores into $k$ evenly distributed bins, which allows us
  to use a bucket priority queue for sorting. This way, the agglomeration can
  be carried out with a worst-case linear runtime.

% Summary, praise, related work
%
The resulting method (prediction of affinities, watershed, and agglomeration)
scales favourably with $O(n)$ in the size $n$ of the volume, a crucial property
for neuron segmentation from EM volumes, where volumes easily reach several
hundreds of terabytes.
  This is a major advantage over current state-of-the-art
  methods that all follow a similar pattern. First, voxel-wise predictions are
  made using a deep neural network. Subsequently, fragments are obtained from
  these predictions which are then merged using either greedy
  (\celis~\cite{Maitin-Shepard2016}, \gala~\cite{Nunez2014}) or globally
  optimal objectives (\multicut~\cite{Andres2012} and lifted
  \multicut~\cite{Keuper2015,Beier2017}).
  {\rerevised
  All these methods depend heavily on the quality of the initial fragments,
  which in turn depend on the quality of the boundary prediction. Despite this
  strong coupling, the boundary classifier is mostly trained unaware of the
  algorithm used to subsequently extract fragments. A noteworthy exception is a
  recent work~\cite{Wolf2017} where a boundary classifier is trained using a
  structured loss to fill objects with seeded watershed regions. This work
  demonstrates the usefulness of structured boundary prediction, similar in
  spirit to the method described here.
  Nevertheless, the majority of current efforts focuses on the merging of
  fragments: Both \celis and \gala train a classifier to predict scores for
  hierarchical agglomeration which increases the computational complexity of
  agglomeration during inference. Similarly, the \multicut variants train a
  classifier to predict the connectivity of fragments that are then clustered
  by solving a computationally expensive combinatorial optimization problem.
  }
  Our proposed fragment agglomeration method drastically reduces the
  computation complexity compared to previous merge methods and does not
  require a separate training step.

% Technical notes
%
%All our software components are based on open source software and made publicly
%available\footnote{\url{https://anonymized}}.
%%
  %We implemented the 3D \unet and \constmalis in \caffe, which allows efficient
  %training and processing of volumes on consumer hardware\footnote{We used
  %nVidia's GeForce Titan X with 12 GB RAM.}.
  %%
  %We implemented our agglomeration method in a highly configurable C++ template
  %library, which provides a Python wrapper (with on-the-fly compilation) for
  %easy integration into existing workflows.

% Preview of results
%
We demonstrate the efficacy of our method on three diverse datasets of EM
volumes, imaged by three different 3D electron microscopy techniques: \cremi (ssTEM, \emph{Drosophila}), \fib (FIBSEM, \emph{Drosophila}),
and \segem (SBEM, mouse cortex).
  Our method significantly improves over the current state of the art in each
  of these datasets, outperforming in particular computationally more expensive
  methods without favorable worst-case runtime guarantees.

We made the source code for
training\footnote{\url{https://github.com/naibaf7/caffe}} and
agglomeration\footnote{\url{https://github.com/funkey/waterz}} publicly
available, together with usage example scripts to reproduce our \cremi
results\footnote{\url{http://cremi.org/static/data/20170312_mala_v2.tar.gz}}.

\section{Method}

\subsection{Deep multi-scale convolutional network for predicting 3D voxel affinities}

\begin{figure*}[h]
  %\begin{subfigure}[t]{\textwidth}
    \centerline{\scalebox{0.9}{\includegraphics{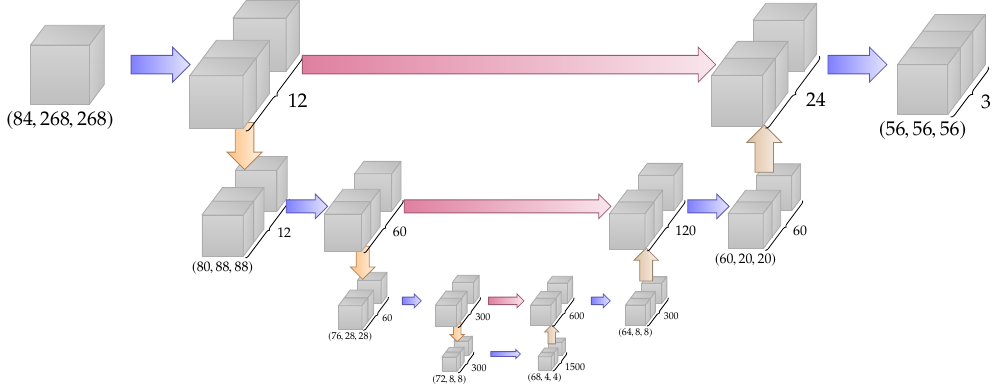}}}
    \caption{Overview of the U-net architecture used for the \cremi dataset.
    The architectures for \fib and \segem are similar, with changes in the
    input and output sizes (in: $(132,132,132)$, out: $(44,44,44)$ for \fib and
    in: $(188,188,144)$, out: $(100,100,96)$ for \segem) and number of feature
    maps for \fib (24 in the first layer, increased by a factor of 3 for lower
    layers).}
  %\end{subfigure}
    \label{fig:supplemental:unet:architecture}
\end{figure*}

\begin{figure*}
  %\begin{subfigure}{\textwidth}
    \centerline{\scalebox{0.9}{\includegraphics{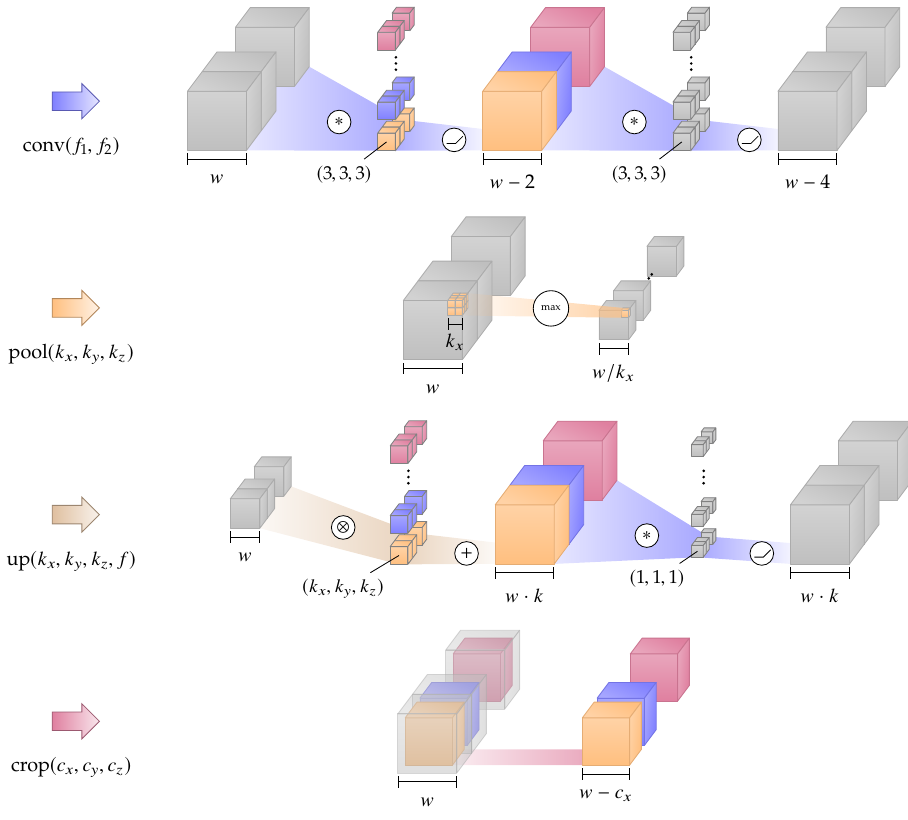}}}
    \caption[]{Details of the convolution (blue), max-pooling (yellow),
    upsampling (brown), and copy-crop operations (red). ``$*$'' denotes a
    convolution,
    ``\tikz[scale=0.25]{\draw(-0.9,-0.25)--(0.1,-0.25)--(0.6,0.25);}'' a
    rectified linear unit, and ``$\otimes$'' the Kronecker matrix product.}
    \label{fig:supplemental:unet:passes}
  %\end{subfigure}
\end{figure*}

% What architecture are we using?
%
We use a 3D \unet architecture~\cite{Cicek2016} to predict voxel affinities on
3D volumes.
  We use the same architecture for all investigated datasets which we
  illustrate in \figref{fig:method:overview:unet}. In particular, our 3D \unet
  consists of four levels of different resolutions. In each level, we perform
  at least one convolution pass (shown as blue arrows in
  \figref{fig:method:overview:unet}) consisting of two convolutions (kernel
  size $3\x3\x3$) followed by rectified linear units.
  Between the levels, we perform max pooling on variable kernel sizes depending
  on the dataset resolution for the downsampling pass (yellow arrows), as well
  as transposed convolution of the same size for upsampling (brown
  arrows).
  The results of the upsampling pass are further concatenated with copies of
  the feature maps of the same level in the downsampling pass (red arrows),
  cropped to account for context loss in the lower levels. Details of the individual passes are
  shown in \figref{fig:supplemental:unet:passes}.
  A more detailed description of the \unet architectures for each of the
  investigated datasets can be found in
  \figref{fig:supplemental:unet:architecture}.

% Why affinities?
%
We chose to predict voxel affinities on edges between voxels instead of
labeling voxels as foreground/background to allow our method to handle low
spatial resolutions.
  As we illustrate in \figref{fig:method:overview:affinities}, a low z
  resolution (common for serial section EM) renders a foreground/background labeling of voxels
  impossible. Affinities, on the other hand, effectively increase the
  expressiveness of our model and allow to obtain a correct segmentation.
  Furthermore, affinities easily generalize to arbitrary neighborhoods and
  might thus allow the prediction of longer range connectivity.

\subsection{Training using \constmalis}

\begin{figure*}[t]
  \centerline{
    \begin{subfigure}[t]{0.25\textwidth}
      \centerline{\includegraphics[width=\textwidth]{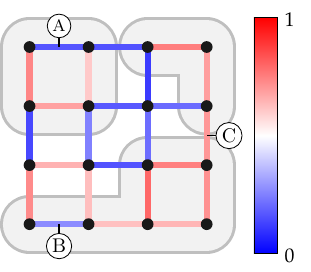}}
      \caption{Predicted affinities.}
      \label{fig:method:malis:affs_gt}
    \end{subfigure}
    \begin{subfigure}[t]{0.25\textwidth}
      \centerline{\hspace{4mm}\includegraphics[width=\textwidth]{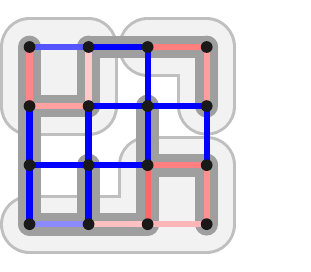}\hspace{-4mm}}
      \caption{Positive pass.}
      \label{fig:method:malis:pos}
    \end{subfigure}
    \begin{subfigure}[t]{0.25\textwidth}
      \centerline{\hspace{4mm}\includegraphics[width=\textwidth]{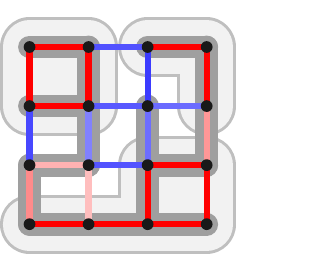}\hspace{-4mm}}
      \caption{Negative pass.}
      \label{fig:method:malis:neg}
    \end{subfigure}
    \begin{subfigure}[t]{0.25\textwidth}
      \centerline{\includegraphics[width=\textwidth]{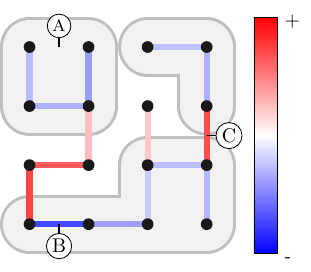}}
      \caption{Gradient of loss.}
      \label{fig:method:malis:loss}
    \end{subfigure}
  }
  \caption{Illustration of the \constmalis{}loss. Given predicted affinities
  (blue low, red high) and a ground-truth segmentation
  \subfigref{fig:method:malis:affs_gt}, losses on maximin edges are computed in
  two passes: In the positive pass, \subfigref{fig:method:malis:pos}, affinities
  of edges between ground-truth regions are set to zero (blue), in the negative
  pass \subfigref{fig:method:malis:neg}, affinities within ground-truth regions
  are set to one (red). In either case, a maximal spanning tree
  (shown as shadow) is constructed to identify maximin edges. Note that,
  in this example, edge A is not a maximin edge in the positive pass since the
  incident voxels are already connected by a high affinity path. In contrast,
  edge B is the maximin edge of the bottom left voxel to any other voxel in the
  same region and thus contributes to the loss. Similarly, C is the maximin
  edge connecting voxels of different ground-truth regions and contributes
  during the negative pass to the loss. The resulting gradients of the loss
  with respect to each edge affinity is shown in
  \subfigref{fig:method:malis:loss} (positive values in red, negative in
  blue).} \label{fig:method:malis}
\end{figure*}

% What is MALIS?
%
We train our network using an extension of the \malis loss~\cite{Turaga2009}.
  This loss, that we term \emph{constrained} \malis, is designed to minimize topological
  errors in a segmentation obtained by thresholding and connected component
  analysis. Although thresholding alone will unlikely produce accurate results,
  it serves as a valuable proxy for training: If the loss can be minimized for
  thresholding, it will in particular be minimized for agglomeration.
  To this end, in each training iteration, a complete affinity prediction of a
  3D region is considered. Between every pair of voxels, we determine the
  maximin affinity edge, \ie, the highest minimal edge over all paths
  connecting the pair. This edge is crucial as it determines the threshold
  under which the two voxels in question will be merged. Naturally, for voxels
  that are supposed to belong to the same region, we want the maximin edge
  affinity to be as high as possible, and for voxels of different regions as
  low as possible.

% What is our extension?
%
Our extension consists of two parts:
  {\rerevised First, we improve the computational complexity of the \malis loss
  by presenting an $O(n\log(n) + kn)$ method for the computation of the
  gradient, where $n$ is the size of the volume and $k$ the number of
  ground-truth objects. We thus improve over the previous method that had a
  complexity of $O(n^2)$}. Second, we compute the gradient in two separate
  passes, once for affinities inside ground-truth objects (positive pass), and
  once for affinities between and outside of ground-truth objects.

{\revised
\subsubsection{The MALIS loss}
}
%
% How to find the maximin edges efficiently?
%
{\revised
Let $G=(V,E,\aff)$ be an affinity graph on voxels $V$ with edges $E\subseteq
V^2$ and affinities $\aff: E \mapsto [0,1]$.
}
A maximin edge between two voxels $u$ and $v$ is an edge $\mm(u,v)\in E$ with
lowest affinity on the overall highest affinity path $P^*_{u,v}$ connecting $u$
and $v$, \ie,
  \begin{equation}
    P^*_{u,v} = \argmax{P\in\mathcal{P}_{u,v}} \min_{e \in P} \aff(e)
    \;\;\;\;\;\;\;\;
    \mm(u,v) = \argmin{e \in P^*_{u,v}} \aff(e)
    \text{,}
    \label{eq:method:maximin}
  \end{equation}
  where $\mathcal{P}_{u,v}$ denotes the set of all paths between $u$ and $v$.
  If we imagine a simple thresholding on the affinity graph, such that edges
  with affinities below a threshold $\theta$ are removed {\revised from $G$},
  then the affinity of the maximin edge $\mm(u,v)$ is equal to the highest
  threshold under which nodes $u$ and $v$ would still be part of the same
  connected component.
  {\revised
  Acknowledging the importance of maximin edges, the \malis loss favors high
  maximin affinities between voxels that belong to the same ground-truth
  segment, and low maximin affinities between voxels that belong to different
  ground-truth segments.
  We assume that a ground-truth segmentation is given as a labelling $\seg:V
  \mapsto \{0,...,k\}$ such that each segment has a unique label in
  $\{1,...,k\}$ and background is marked with $0$. Let $F\subseteq V$ denote
  all foreground voxels $F = \{ v\in V \;|\; \seg(v)\neq0 \}$ and $\delta(u,
  v)$ indicate whether $u$ and $v$ belong to the same ground-truth segment:
  \begin{equation}
    \delta(u, v) = \left\{
      \begin{array}{ll}
        1 & \text{if}\; u,v \in F \;\text{and}\; \seg(u) = \seg(v)\text{,} \\
        0 & \text{otherwise.} \\
      \end{array}
      \right.
  \end{equation}
  The \malis loss $\malisloss(\seg, \aff)$ is the sum of affinity losses over
  the maximin edges of every pair of voxels that do not belong to the
  background:
  \begin{equation}
    \malisloss(\seg, \aff) = \sum_{u,v\in F} \affloss\left(\delta(u, v), \aff(\mm(u,v))\right)
    \text{.}
    \label{eq:method:malis_loss}
  \end{equation}
  The affinity loss can be any continuous and differentiable function, we chose
  $\affloss(x, y) = (x - y)^2$ for all experiments in this paper.

\vspace{4mm}% why is this needed here?
\subsubsection{Quasilinear loss computation}
Considering that we have $O(n^2),\;n=|V|$, pairs of voxels, but---in the case
of grid graphs considered here---only $O(n)$ edges, it follows that maximin
edges are shared between voxel pairs.
  This observation generalizes to arbitrary graphs. In particular, the union of
  all maximin edges forms a maximal spanning tree (MST),
  \begin{equation}
    \left\{ \mm(u,v) \;|\; (u,v) \in V^2 \right\} = \text{MST}(G)
    \text{,}
  \end{equation}
  \ie, there are always only $n-1$ maximin edges in a graph.

That the previous equality holds can easily be proven by contradiction:
  Assume that for a pair $(u,v)$, $\mm(u,v)\notin\text{MST}(G)$. Let
  $P_{u,v}^+\subseteq \text{MST}(G)$ denote the path connecting $u$ and $v$ on the
  MST, and let $\mtp(u,v)$ denote the edge with minimal affinity on
  $P_{u,v}^+$:
  \begin{equation}
    \mtp(u,v) = \argmin{e \in P_{u,v}^+} \aff(e)
    \text{.}
  \end{equation}
  Following our assumption, $P_{u,v}^+$ does not contain $\mm(u,v)$. By
  definition \eqref{eq:method:maximin}, the following inequalities hold:
  \begin{equation}
    \aff(\mtp(u,v)) \leq \aff(\mm(u,v)) \leq \aff(e) \;\;\forall e \in P_{u,v}^*
    \text{.}
  \end{equation}
  We can now remove $\mtp(u,v)$ from the MST to obtain two disconnected
  sub-trees separating $u$ from $v$. Since $P_{u,v}^*$ connects $u$ and $v$,
  there exists an edge $e^*\in P_{u,v}^*$ that will reconnect the two
  sub-trees. However, $\aff(\mtp(u,v)) \leq \aff(e^*)$. If strict inequality
  holds, this will create a tree with a larger sum of affinities than the MST,
  thus contradicting our assumptions. If equality holds and $\aff(\mtp(u,v)) =
  \aff(\mm(u,v)) = \aff(e^*) $, then there are more than one possible MSTs and
  hence $\mm(u,v)$ is contained in one of them.
  }

Consequently, we are able to identify the maximin edge and compute its loss for
each voxel pair by growing an MST on $G$.
  {\revised
  We use Kruskal's algorithm~\cite{Kruskal1956} to grow an MST{\rerevised,
  which consists of two steps: First, we sort all edges by affinity in
  descending order. Second, we iterate over all edges and grow the MST using a
  union-find data structure.} Whenever a new edge $e$ merges two trees $T_1,
  T_2 \subset MST(G)$ during construction of the MST, we compute the
  \emph{positive} and \emph{negative} weight of {\rerevised this} edge on the
  fly. The positive weight $\weight_P(e)$ corresponds to the number of voxel
  pairs of the same ground-truth segment merged by $e$:
  \begin{equation}
    \weight_P(e) = |\{ (u,v) \in F^2 \;|\; \delta(u,v)=1,\; e=\mm(u,v) \}|
    \text{.}
  \end{equation}
  By construction, $e$ is the maximin edge to all pairs of voxels between the
  two trees it merges. Therefore, $\weight_P(e)$ equals the product of the
  number of voxels having label $i$ in either tree, summed over all
  $i\in\{1,\ldots,k\}$. {\rerevised Let $V_T$ denote the set of voxels in $T$
  and $V_T^i\subseteq V_T$ the subset with ground-truth label $i$.} The
  positive weight can {\rerevised then be rewritten} as:
  \begin{equation}
    \weight_P(e) = \sum_{i\in\{1,\ldots,k\}} \left|V_{T_1}^i\right|\left|V_{T_2}^i\right|
    \label{eq:method:weight_pos}
    \text{.}
  \end{equation}
  Equivalently, the negative weight $\weight_N(e)$ is the number of voxel pairs
  of different ground-truth segments merged by $e$:
  {\rerevised
  \begin{align}
    \weight_N(e) &= |\{ (u,v) \in F^2 \;|\; \delta(u,v)=0,\; e=\mm(u,v) \}| \\
                 &= \sum_{i\neq j\in\{1,\ldots,k\}} \left|V_{T_1}^i\right|\left|V_{T_2}^j\right| \\
                 &= \left|V_{T_1}\right|\left|V_{T_2}\right| - \sum_{i\in\{1,\ldots,k\}} \left|V_{T_1}^i\right|\left|V_{T_2}^i\right|
    \label{eq:method:weight_neg}
    \text{.}
  \end{align}
  }
  We can now rewrite the \malis loss~\eqref{eq:method:malis_loss} as
  \begin{equation}
    L(\seg,\aff) = \sum_{e\in \text{MST}(G)} \weight_P(e)\affloss(1, \aff(e)) + \weight_N(e)\affloss(0, \aff(e))
  \end{equation}
  and avoid the costly sum over all pairs of voxels.
  {\rerevised We keep track of the sizes of sets $V_T$ and $V_T^i$ used in each
  tree during the construction of the MST. Consequently, the complexity of our
  algorithm is dominated by first sorting all edges by their affinity in
  $O(n\log(n))$ and subsequently evaluating
  equations~\eqref{eq:method:weight_pos} and~\eqref{eq:method:weight_neg} while
  constructing the MST in $O(kn)$, resulting in a final complexity of
  $O(n\log(n)+kn)$.}}
  We thus improve over a previous method~\cite{Turaga2009} that required
  $O(n^2)$ and therefore had to fall back to sparse sampling of voxel pairs.
  Note that this only affects the training of the network, the affinity
  prediction during test time scales linearly with the volume size.

{\revised
\subsubsection{Constrained MALIS}
}
%
% What is the point of the 2 passes?
%
We further extend previous work by computing the maximin edge losses in two
passes:
  {\rerevised In the first pass we compute only the weights $\weight_P$ for
  edges within the same region (positive pass). In the second pass, we compute
  the weights $\weight_N$ for edges between different regions (negative
  pass).} As shown in \figref{fig:method:malis}, in the positive pass, we
  assume that all edges between regions have been predicted correctly and set
  their affinities to zero. Consequently, only maximin edges inside a region
  are found and contribute to the loss. This obviates an inefficiency in a
  previous formulation~\cite{Turaga2009}, where a spurious high-affinity (\ie,
  false positive) path leaving and entering a region might connect two voxels
  inside the same region. In this case, the maximin edge could lie outside of
  the considered region, resulting in an unwanted gradient contribution that
  would reinforce the false positive. Analogously, in the negative pass, all
  affinities inside the same region are set to one to avoid reinforcement of
  false negatives inside regions. Finally, the gradient contributions of both
  passes are added together.

{\rerevised
Note that, similar to the original \malis formulation~\cite{Turaga2009}, the
constrained version presented here does not require precise location of the
boundaries.
  In applications where the exact location of the boundary is less relevant, a
  broader background region around boundaries can be given. During the negative
  pass, any correctly predicted cut through this background region will result
  in a loss of zero.
}

\subsection{Hierarchical agglomeration}

\begin{figure}[t]
  \centerline{\subimport{../figures/}{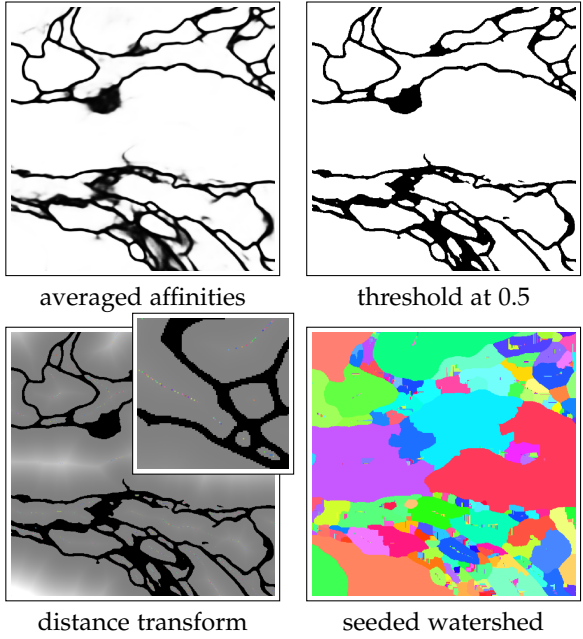}}
  \caption{Illustration of the seeded watershed heuristic.}
  \label{fig:method:watershed}
\end{figure}

Our method for hierarchical agglomeration of segments from the predicted
affinities consists of two steps.
  First, we use a heuristic to extract small fragments directly from the
  predicted affinities. Second, we iteratively score and merge adjacent
  fragments into larger objects until a predefined threshold is reached.

\subsubsection{Fragment extraction}

% Why fragments at all?
%
The extraction of fragments is a crucial step for the subsequent agglomeration.
  Too many fragments slow down the agglomeration unnecessarily and increase its
  memory footprint. Too few fragments, on the other hand, are subject to
  undersegmentation that cannot be corrected.

% How do we get them?
%
Empirically, we found a seeded watershed to deliver the best trade-off between
fragment size and segmentation accuracy across all investigated datasets.
  For the seeded watershed, we first average the predicted affinities for each
  voxel to obtain a volume of boundary predictions. We subsequently threshold
  the boundary predictions at $0.5$ and perform a distance transform on the
  resulting mask. Every local maximum is taken as a seed, from which we grow
  basins using a standard watershed algorithm~\cite{Coelho2013} on the boundary
  predictions. For an example, see \figref{fig:method:watershed}.
  As argued above, voxel-wise predictions are not fit for anisotropic volumes
  with low z-resolution (see \figref{fig:method:overview:affinities}). To not
  re-introduce a flaw that we aimed to avoid by predicting affinities instead
  of voxel-wise labels in the first place, we perform the extraction of
  fragments xy-section-wise for anisotropic volumes.

\subsubsection{Fragment agglomeration}
\label{sec:method:agglomeration:agglomeration}

% How does agglomeration work in general?
%
For the agglomeration, we consider the region adjacency graph (RAG) of the
extracted fragments.
  The RAG is an annotated graph $G=(V,E,f)$, with $V$ the set of fragments, $E
  \subseteq V\times V$ edges between adjacent fragments, and $f: E \mapsto
  \mathbb{R}$ an edge scoring function. The edge scoring function is designed
  to prioritize merge operations in the RAG, \ie, the contraction of two
  adjacent nodes into one, such that edges with lower scores are merged
  earlier. Given an annotated RAG, a segmentation can be obtained by finding
  the edge with the lowest score, merge it, recompute the scores of edges
  affected by the merge, and iterate until the score of the lowest edge hits a
  predefined threshold $\theta$. In the following, we will denote by $G_i$ the
  RAG after $i$ iterations (and analogously by $V_i$, $E_i$, and $f_i$ its
  nodes, edges, and scores), with $G_0 = G$ as introduced above. We will
  "reuse" nodes and edges, meaning $V_{i+1} \subset V_i$ and $E_{i+1} \subset
  E_i$.

% What is the role of the merge function?
%
Given that the initial fragments are indeed an oversegmentation, it is up to
the design of the scoring function and the threshold $\theta$ to ensure a
correct segmentation.
  \begin{figure}[t]
    \centerline{\begin{tikzpicture}

  \begin{scope}[rotate=40,scale=0.9]
  \node[region] (r1) at (0,0) {D};
  \node[region] (r2) at (2,-0.5) {A};
  \node[region] (r3) at (1.5,-1.5) {B};
  \node[region] (r5) at (3.5,-2) {C};
  \node[region] (r4) at (4,0) {E};
  \coordinate (left_center) at (1.75,-1);
  \end{scope}
  \draw (r1) -- node[pos=0.5,above] {d} (r2);
  \draw (r1) -- node[pos=0.5,below] {c} (r3);
  \draw (r2) -- node[pos=0.5,left]  {} (r3);
  \draw (r2) -- node[pos=0.5,above] {e} (r4);
  \draw (r3) -- node[pos=0.5,below] {b} (r5);
  \draw (r4) -- node[pos=0.5,right] {a} (r5);
  \begin{pgfonlayer}{background}
  \draw[merge_highlight] (r3.center)--(r2.center);
  \end{pgfonlayer}

  \begin{scope}[yshift=-3cm,xshift=3cm]
    \begin{scope}[rotate=40,scale=0.9]
    \node[region] (r12) at (0,0) {D};
    \node[region] (r22) at (1.75,-1) {A};
    \node[region] (r52) at (3.5,-2) {C};
    \node[region] (r42) at (4,0) {E};
    \coordinate (right_center) at (1.75,-1);
    \end{scope}
    \draw (r12) -- node[pos=0.5,below] {c} (r22);
    \draw (r22) -- node[pos=0.5,above] {e} (r42);
    \draw (r22) -- node[pos=0.5,below] {b} (r52);
    \draw (r42) -- node[pos=0.5,right] {a} (r52);
  \end{scope}

  \path (left_center) -- node[process,pos=0.55,rotate=-45] {merge A,B$\rightarrow$A} (right_center);

\end{tikzpicture}}
    \caption{\rerevised Illustration of the three different edge update cases
    during a merge. Case 1: The edge is not involved in the merge at all ($a$).
    Case 2: One of the edge's nodes is involved in the merge, but the boundary
    represented by the edge does not change ($b$ and $e$). Case 3: The
    boundaries represented by two edges get merged ($c$ and $d$). Only in this
    case the score needs to be updated.} \label{fig:method:score_update}
  \end{figure}
  The design of the scoring function can be broken down into the initialization
  of $f_0(e)$ for $e \in E_0$ (\ie, the initial scores) and the update of
  $f_i(e)$ for $e \in E_i;\;i>0$ after a merge of two regions $a, b \in
  V_{i-1}$. For the update, three cases can be distinguished (for an
  illustration see \figref{fig:method:score_update}): (1) $e$ was not affected
  by the merge, (2) $e$ is incident to $a$ or $b$ but represents the same
  contact area between two regions as before, and (3) $e$ results from merging
  two edges of $E_{i-1}$ into one (the other edge will be deleted). In the
  first two cases, the score does not change, \ie, $f_i(e) = f_{i-1}(e)$, since
  the contact area between the nodes linked by $e$ remains the same. In the
  latter case, the contact area is the union of the contact area of the merged
  edges, and the score needs to be updated accordingly. Acknowledging the merge
  hierarchy of edges (as opposed to nodes), we will refer to the leaves under a
  merged edge $e$ as \emph{initial edges}, denoted by $E^*(e) \subseteq E_0$.

% What merge function(s) are we considering?
%
In our experiments, we initialize the edge scores $f(e)$ for $e \in E_0$ with
one minus the maximum affinity between the fragments linked by $e$ and update
them using a quantile value of scores of the initial edges under $e$.
  This strategy has been found empirically over a range of possible
  implementations of $f$ (see \secref{sec:results:mergefunction}).

% What is the runtime complexity of agglomeration?
%
Implemented naively, hierarchical agglomeration has a worst-case runtime
complexity of at least $O(n\log(n))$, where $n=|E_0|$ is the number of edges in
the initial RAG.
  This is due to the requirement of finding, in each iteration, the cheapest
  edge to merge, which implies sorting of edges based on their scores.
  Furthermore, the edge scoring function has to be evaluated $O(n)$ times, once
  for each affected edge of a node merge (assuming nodes have a degree bounded
  by a constant). For the merge function suggested above, a quantile of $O(n)$
  initial edge scores has to be found in the worst case\jantodo{would a
  balanced merge tree improve this figure to $O(\log(n))$?}, resulting in a
  total worst-case runtime complexity of $O(n\log(n) + n^2)$.

% How do we get linear runtime?
%
To avoid this prohibitively high runtime complexity, we propose to discretize
the initial scores $f_0$ into $k$ bins, evenly spaced in the interval $[0,1]$.
  This simple modification has two important consequences: First, a bucket
  priority queue for sorting edge scores can be used, providing constant time
  insert and pop operations. Second, the computation of quantiles can be
  implemented in constant time and space by using histograms of the $k$
  possible values. This way, we obtain constant-time merge iterations (pop an
  edge, merge nodes, update scores of affected edges), applied at most $n$
  times, thus resulting in an overall worst-case complexity of $O(n)$.
  With $k=256$ bins, we noticed no sacrifice of accuracy in comparison to the
  non-discretized variant.

% How else do we speed up agglomeration?
%
The analysis above holds only if we can ensure that the update of the score of
an edge $e$, and thus the update of the priority queue, can be performed in
constant time.
  In particular, it is to be avoided to search for $e$ in its respective
  bucket. We note that for the quantile scoring function (and many more), the
  new edge score $f_i(e)$ after merging an edge $f\in E_{i-1}$ into $e\in
  E_{i-1}$ is always greater than or equal to its previous score. We can therefore mark $e$
  as \emph{stale} and $f$ as \emph{deleted} and proceed merging without
  resorting the queue or altering the graph. Whenever a stale edge is popped
  from the priority queue, we compute its actual score and insert it again into
  the queue. Not only does this ensure constant time updates of edge scores and
  the priority queue, it also avoids computing scores for edges that are never
  used for a merge. This can happen if the threshold is hit before considering
  the edge, or if the edge got marked as deleted as a consequence of a nearby
  merge.

\section{Results}

% U-Net Overview figure
%

% U-Net details figure
%

% Main results figure
%
\begin{table}[t]%
  \begin{subtable}{0.5\textwidth}
    \centerline{\rowcolors{2}{gray!2!white}{gray!20!white}
\begin{tabular}{l|D{.}{.}{3}D{.}{.}{3}D{.}{.}{3}}
    method&\multicolumn{1}{c}{VOI split}&\multicolumn{1}{c}{VOI merge}&\multicolumn{1}{c}{VOI sum}\\

    \hline
\bf \unet \mala&0.891&0.180&\multicolumn{1}{B{.}{.}{3} }{1.071}\\
\bf \unet&1.205&0.316&1.520\\
FlyEM~\cite{Takemura2015}&1.490&0.462&1.952\\
CELIS~\cite{Maitin-Shepard2016}&1.426&0.208&1.634\\
CELIS+MC~\cite{Maitin-Shepard2016}&1.037&0.229&1.266
\end{tabular}
}
    \caption{Results on \fib, evaluated on whole test volume.}
    \label{tab:results:fib:volumetric}
  \end{subtable}
  \begin{subtable}{0.5\textwidth}
    \vspace{5.7mm}
    \centerline{\rowcolors{2}{gray!2!white}{gray!20!white}
\begin{tabular}{l|D{.}{.}{3}D{.}{.}{3}D{.}{.}{3}}
    method&\multicolumn{1}{c}{VOI split}&\multicolumn{1}{c}{VOI merge}&\multicolumn{1}{c}{VOI sum}\\

    \hline
\bf \unet \mala&1.953&0.198&\multicolumn{1}{B{.}{.}{3} }{2.151}\\
\bf \unet&2.442&0.471&2.914\\
FlyEM~\cite{Takemura2015}&3.160&0.251&3.411\\
CELIS~\cite{Maitin-Shepard2016}&3.401&0.166&3.568\\
CELIS+MC~\cite{Maitin-Shepard2016}&2.354&0.216&2.570
\end{tabular}
}
    \caption{Results on \fib, evaluated on synaptic sites.}
    \label{tab:results:fib:synapse}
  \end{subtable}
  \begin{subtable}{0.5\textwidth}
    \vspace{5.7mm}
    \centerline{\scalebox{0.9}{\rowcolors{2}{gray!2!white}{gray!20!white}
\begin{tabular}{l|D{.}{.}{3}D{.}{.}{3}D{.}{.}{3}D{.}{.}{3}}
    method&\multicolumn{1}{c}{VOI split}&\multicolumn{1}{c}{VOI merge}&\multicolumn{1}{c}{VOI sum}&\multicolumn{1}{c}{CREMI score}\\

    \hline
\bf \unet \mala&0.425&0.181&\multicolumn{1}{B{.}{.}{3} }{0.606}&\multicolumn{1}{B{.}{.}{3} }{0.289}\\
\bf \unet&0.979&0.546&1.524&0.793\\
LMC~\cite{Beier2017}&0.597&0.272&0.868&0.398\\
CRunet~\cite{Zeng2017}&1.081&0.389&1.470&0.566\\
LFC~\cite{Parag2017}&1.085&0.140&1.225&0.616
\end{tabular}
}}
    \caption{Results on \cremi (from leaderboard in~\cite{Cremi}).}
    \label{tab:results:cremi}
  \end{subtable}
  \begin{subtable}{0.5\textwidth}
    \vspace{5.7mm}
    \centerline{\rowcolors{2}{gray!2!white}{gray!20!white}
\begin{tabular}{l|D{.}{.}{3}D{.}{.}{3}D{.}{.}{3}}
    method&\multicolumn{1}{c}{IED split}&\multicolumn{1}{c}{IED merge}&\multicolumn{1}{c}{IED total}\\

    \hline
\bf \unet \mala&6.259&21.337&\multicolumn{1}{B{.}{.}{3} }{4.839}\\
\bf \unet&6.903&1.719&1.377\\
SegEM~\cite{Berning2015}&2.121&3.951&1.380
\end{tabular}
}
    \caption{Results on \segem.}
    \label{tab:results:segem}
  \end{subtable}
  \caption{Qualitative results of our method (\unet \mala) compared to the
  respective state of the art on the testing volumes of each dataset and a
  baseline (\unet). Highlighted in bold are the names of our method and the
  best value in each column.
  Measures shown are variation of information (VOI, lower is better), CREMI
  score (geometric mean of VOI and adapted RAND error, lower is better), and
  inter-error distance in \textmu m (IED, higher is better) evaluated on traced
  skeletons of the test volume. The IED has been computed using the TED metric
  on skeletons~\cite{Funke2017} with a distance threshold of 52\,nm
  (corresponding to the thickness of two z-sections).
  \cremi results are reported as average over all testing samples, individual
  results can be found in \figref{fig:supplemental:results:cremi}.
  }
  \label{tab:results}
\end{table}

\begin{figure*}[t]%
  \def\plotheight{5cm}
  \begin{subfigure}{0.5\textwidth}
    \centerline{
      \def\plotwidth{0.9\textwidth}
      \begin{tikzpicture}
    \begin{axis}[
        ymajorgrids=true,
        xmajorgrids=true,
        width=\plotwidth,
        height=\plotheight,
        legend style={font=\tiny},
        legend image post style={scale=0.5},
        legend columns=1,
        xlabel=VOI merge,
        ylabel=VOI split,
        xticklabel style={font=\tiny,overlay},
        yticklabel style={font=\tiny,overlay},
        ylabel style={font=\tiny,overlay},
        xlabel style={font=\tiny,overlay},,xmin=0.000000,xmax=1.000000,ymin=0.000000,ymax=2.000000
    ]
        
    \addplot[,red!75!black]
        coordinates {
            (0.005582,8.034401)(0.017698,1.544175)(0.019759,1.344417)(0.022113,1.189945)(0.023310,1.122902)(0.024402,1.049750)(0.025483,1.009739)(0.026579,0.970489)(0.027298,0.942300)(0.028865,0.915870)(0.030235,0.893128)(0.030541,0.882551)(0.032616,0.867161)(0.034015,0.852840)(0.037088,0.832943)(0.037527,0.824469)(0.040633,0.808372)(0.041230,0.795851)(0.041785,0.788225)(0.042432,0.774607)(0.042877,0.768626)(0.043399,0.757717)(0.043815,0.749146)(0.044340,0.738610)(0.045387,0.732756)(0.046024,0.723824)(0.046440,0.715878)(0.047104,0.709837)(0.047753,0.703107)(0.047965,0.698897)(0.048320,0.693291)(0.048775,0.688167)(0.049104,0.682726)(0.049616,0.675022)(0.050246,0.669286)(0.051018,0.664393)(0.051496,0.660262)(0.052295,0.653463)(0.056628,0.648821)(0.058518,0.642496)(0.058896,0.639570)(0.059888,0.633858)(0.060729,0.627282)(0.061179,0.624207)(0.062516,0.617720)(0.062972,0.613849)(0.063465,0.608839)(0.063873,0.606146)(0.065554,0.601483)(0.068416,0.598297)(0.069919,0.593771)(0.071691,0.589157)(0.073280,0.584897)(0.074493,0.581542)(0.075532,0.577686)(0.077615,0.574478)(0.078165,0.571960)(0.081559,0.565576)(0.082671,0.562305)(0.083739,0.558696)(0.084467,0.554469)(0.085295,0.550473)(0.086064,0.544959)(0.086488,0.540968)(0.087739,0.537042)(0.088662,0.534770)(0.091171,0.531254)(0.092636,0.527551)(0.093743,0.524450)(0.095141,0.520075)(0.096186,0.517059)(0.097618,0.512455)(0.098780,0.509331)(0.099789,0.505377)(0.102593,0.502704)(0.114853,0.496992)(0.118165,0.492425)(0.120738,0.489759)(0.128821,0.485290)(0.132368,0.482093)(0.137195,0.474282)(0.140976,0.470332)(0.147059,0.464783)(0.149514,0.460410)(0.163308,0.455664)(0.170624,0.449838)(0.179051,0.444206)(0.193033,0.436334)(0.202784,0.430441)(0.220781,0.420782)(0.236526,0.411810)(0.267675,0.402678)(0.310289,0.390896)(0.356017,0.383368)(0.448924,0.370847)(0.493358,0.360248)(0.621227,0.346307)(0.819560,0.333862)(1.695167,0.263436)(3.380930,0.153539)
        };
    \addlegendentry{ \bf \unet \mala }

    \addplot[,blue!50!white]
        coordinates {
            (0.003072,8.520642)(0.005155,4.790537)(0.009165,3.943182)(0.011233,3.262542)(0.013320,2.999181)(0.018424,2.709330)(0.019423,2.561432)(0.021420,2.389953)(0.024320,2.298899)(0.027210,2.184731)(0.032557,2.092701)(0.035847,2.042665)(0.044470,1.968879)(0.046082,1.929343)(0.049634,1.860415)(0.052832,1.826561)(0.055754,1.786594)(0.058295,1.739804)(0.063739,1.715201)(0.068497,1.677114)(0.070698,1.654282)(0.078493,1.617818)(0.083614,1.594072)(0.090598,1.564262)(0.094299,1.544227)(0.098197,1.519136)(0.104796,1.494609)(0.107278,1.479116)(0.116753,1.452147)(0.123162,1.439294)(0.128326,1.410766)(0.134617,1.398734)(0.142509,1.379038)(0.145690,1.365782)(0.154242,1.344674)(0.166679,1.322306)(0.175121,1.309859)(0.184020,1.286927)(0.191270,1.274237)(0.204851,1.256050)(0.210578,1.245923)(0.224813,1.227138)(0.234729,1.211291)(0.247027,1.200050)(0.266950,1.183055)(0.273581,1.165088)(0.294804,1.148142)(0.305710,1.130281)(0.327477,1.113636)(0.345473,1.102790)(0.360184,1.089468)(0.388877,1.076567)(0.398211,1.065336)(0.443010,1.049415)(0.455684,1.039473)(0.513597,1.011410)(0.535177,0.996568)(0.553926,0.976851)(0.585692,0.964297)(0.611970,0.951016)(0.646209,0.929794)(0.670103,0.913953)(0.708605,0.898721)(0.739695,0.889485)(0.800567,0.875845)(0.831754,0.866997)(0.910435,0.851812)(0.962879,0.831899)(0.984623,0.824925)(1.055711,0.809440)(1.080593,0.802058)(1.141108,0.792596)(1.181891,0.780969)(1.248783,0.769925)(1.271848,0.762682)(1.370414,0.751101)(1.548873,0.724482)(1.604928,0.714234)(1.745018,0.688242)(1.839657,0.663856)(2.007124,0.612543)(2.180625,0.591936)(2.286597,0.577057)(2.394755,0.558540)(2.563517,0.544420)(2.723940,0.524705)(2.798607,0.510936)(3.040995,0.478809)(3.239038,0.459329)(3.701489,0.420561)(3.960427,0.374281)(4.267962,0.302111)(4.636780,0.258158)(4.831538,0.243935)(5.248068,0.202400)(5.623883,0.156789)(6.298598,0.079193)(6.779199,0.025356)(7.247996,0.002406)(7.320537,0.000418)
        };
    \addlegendentry{ \bf \unet }

    \addplot[,only marks,mark=*,black]
        coordinates {
            (0.271704,0.596727)
        };
    \addlegendentry{ LMC~\cite{Beier2017} }

    \addplot[,only marks,mark=*,green!75!black]
        coordinates {
            (0.388978,1.081383)
        };
    \addlegendentry{ CRunet~\cite{Zeng2017} }

    \addplot[,only marks,mark=*,brown]
        coordinates {
            (0.139580,1.085207)
        };
    \addlegendentry{ LFC~\cite{Parag2017} }

    \end{axis}
\end{tikzpicture}
    }
    \vspace{7mm}
    \caption{VOI split/merge on \cremi (lower is better).}
    \label{fig:results:split_merge:cremi}
  \end{subfigure}
  \begin{subfigure}{0.5\textwidth}
    \def\plotwidth{0.9\textwidth}
    \centerline{
      \begin{tikzpicture}
    \begin{axis}[
        ymajorgrids=true,
        xmajorgrids=true,
        width=\plotwidth,
        height=\plotheight,
        legend style={font=\tiny},
        legend image post style={scale=0.5},
        legend columns=1,
        xlabel=VOI merge,
        ylabel=VOI split,
        xticklabel style={font=\tiny,overlay},
        yticklabel style={font=\tiny,overlay},
        ylabel style={font=\tiny,overlay},
        xlabel style={font=\tiny,overlay},,xmin=0.000000,xmax=3.000000,ymin=0.000000,ymax=2.000000
    ]
        
    \addplot[,red!75!black]
        coordinates {
            (0.000500,13.631700)(0.108200,1.074600)(0.119900,1.036900)(0.135700,0.992900)(0.139100,0.976100)(0.145400,0.955800)(0.169800,0.908300)(0.180100,0.891100)(0.213900,0.866900)(0.263500,0.846300)(0.312200,0.808600)(0.403800,0.779900)(0.572700,0.730500)(1.292300,0.633000)(6.626900,0.341000)
        };
    \addlegendentry{ \bf \unet \mala }

    \addplot[,blue!50!white]
        coordinates {
            (0.315500,1.205000)(0.404500,1.158600)(0.781300,1.007000)(4.019800,0.651000)(5.843100,0.494300)(8.501200,0.226700)(9.402900,0.122400)(10.395500,0.013400)
        };
    \addlegendentry{ \bf \unet }

    \addplot[,only marks,mark=*,black]
        coordinates {
            (0.461800,1.489800)
        };
    \addlegendentry{ FlyEM~\cite{Takemura2015} }

    \addplot[,only marks,mark=*,green!75!black]
        coordinates {
            (0.207700,1.426000)(0.207700,1.426000)(0.207700,1.426000)
        };
    \addlegendentry{ CELIS~\cite{Maitin-Shepard2016} }

    \addplot[,only marks,mark=*,brown]
        coordinates {
            (0.003200,10.070700)(0.003200,10.070700)(0.228900,1.037300)
        };
    \addlegendentry{ CELIS+MC~\cite{Maitin-Shepard2016} }

    \end{axis}
\end{tikzpicture}
    }
    \vspace{7mm}
    \caption{VOI split/merge on \fib (lower is better).}
    \label{fig:results:split_merge:fib}
  \end{subfigure}
  \begin{subfigure}{0.5\textwidth}
    \vspace{1cm}
    \def\plotwidth{0.9\textwidth}
    \centerline{
      \begin{tikzpicture}
    \begin{axis}[
        ymajorgrids=true,
        xmajorgrids=true,
        width=\plotwidth,
        height=\plotheight,
        legend style={font=\tiny},
        legend image post style={scale=0.5},
        legend columns=1,
        xlabel=distance between merge $\mu m$,
        ylabel=distance between split $\mu m$,
        xticklabel style={font=\tiny,overlay},
        yticklabel style={font=\tiny,overlay},
        ylabel style={font=\tiny,overlay},
        xlabel style={font=\tiny,overlay},,xmode=log,ymode=log
    ]
        
    \addplot[,red!75!black]
        coordinates {
            (inf,0.586031)(36.108528,2.214202)(33.529347,2.301034)(33.529347,2.407235)(31.294058,2.483655)(31.294058,2.622407)(31.294058,2.682348)(31.294058,2.745093)(29.338179,2.761240)(29.338179,2.777579)(29.338179,2.844914)(29.338179,2.897598)(29.338179,2.897598)(29.338179,2.915595)(29.338179,2.952270)(29.338179,2.952270)(29.338179,2.989878)(29.338179,3.028457)(29.338179,3.048122)(29.338179,3.068045)(29.338179,3.068045)(29.338179,3.129406)(29.338179,3.171695)(29.338179,3.215143)(29.338179,3.215143)(29.338179,3.215143)(29.338179,3.237316)(29.338179,3.259798)(29.338179,3.259798)(29.338179,3.259798)(29.338179,3.282593)(29.338179,3.305710)(29.338179,3.329155)(29.338179,3.329155)(29.338179,3.329155)(29.338179,3.329155)(29.338179,3.329155)(29.338179,3.401528)(29.338179,3.401528)(29.338179,3.401528)(29.338179,3.401528)(29.338179,3.426357)(29.338179,3.426357)(29.338179,3.426357)(29.338179,3.426357)(29.338179,3.426357)(29.338179,3.477118)(29.338179,3.477118)(29.338179,3.477118)(29.338179,3.529405)(31.294058,3.556143)(31.294058,3.583289)(31.294058,3.610853)(27.612404,3.667272)(26.078381,3.696149)(26.078381,3.696149)(26.078381,3.696149)(26.078381,3.725483)(26.078381,3.755287)(26.078381,3.755287)(26.078381,3.816348)(26.078381,3.816348)(26.078381,3.816348)(23.470543,3.816348)(23.470543,3.847630)(23.470543,3.847630)(23.470543,3.911757)(23.470543,3.911757)(23.470543,3.944629)(23.470543,4.012059)(23.470543,4.046645)(23.470543,4.081834)(23.470543,4.081834)(23.470543,4.081834)(23.470543,4.117639)(23.470543,4.154078)(23.470543,4.191168)(23.470543,4.267371)(23.470543,4.306522)(23.470543,4.306522)(23.470543,4.306522)(23.470543,4.346397)(23.470543,4.346397)(23.470543,4.387017)(23.470543,4.470580)(23.470543,4.470580)(22.352898,4.557387)(22.352898,4.647632)(22.352898,4.694109)(23.470543,4.789907)(24.705835,4.839287)(24.705835,4.941167)(24.705835,4.941167)(24.705835,5.047429)(24.705835,5.158361)(24.705835,5.215676)(24.705835,5.395527)(24.705835,5.588225)(21.336857,6.258812)(18.776435,6.705869)
        };
    \addlegendentry{ \unet MALA }

    \addplot[,blue!50!white]
        coordinates {
            (inf,0.589712)(156.470288,0.752261)(52.156763,0.838234)(33.529347,0.916818)(26.078381,0.946393)(21.336857,0.994515)(21.336857,1.018245)(18.776435,1.062016)(16.764674,1.107101)(15.647029,1.136588)(15.142286,1.161908)(14.669089,1.179424)(14.224572,1.200539)(13.806202,1.219249)(13.411739,1.241828)(13.039191,1.251762)(12.352917,1.268678)(11.735272,1.296715)(11.449045,1.326019)(11.176449,1.337353)(11.176449,1.341174)(10.916532,1.360611)(10.916532,1.376571)(10.916532,1.401226)(10.916532,1.418160)(10.916532,1.444341)(10.204584,1.480791)(10.204584,1.504522)(10.204584,1.519129)(9.987465,1.524061)(9.579814,1.559505)(9.579814,1.564703)(9.388217,1.585848)(9.579814,1.591223)(9.579814,1.607571)(9.027132,1.624259)(8.534743,1.635578)(8.382337,1.635578)(8.382337,1.641297)(8.093291,1.676467)(8.093291,1.694624)(7.956116,1.725775)(7.956116,1.758093)(7.823514,1.778071)(7.823514,1.791645)(7.823514,1.798509)(7.695260,1.833636)(7.571143,1.855379)(7.695260,1.870163)(7.695260,1.885184)(7.112286,1.908174)(6.803056,1.972315)(6.705869,1.989029)(6.705869,2.040917)(6.611421,2.049829)(6.430286,2.077039)(6.258812,2.104981)(6.176459,2.143429)(6.096245,2.153261)(6.096245,2.183306)(5.941910,2.224696)(5.941910,2.235290)(5.655553,2.289809)(5.395527,2.289809)(5.274279,2.394953)(5.274279,2.419644)(5.215676,2.432181)(5.158361,2.457648)(5.158361,2.496866)(4.889696,2.551146)(4.839287,2.551146)(4.789907,2.579181)(4.602067,2.637140)(4.387017,2.667107)(4.387017,2.697764)(4.306522,2.810843)(4.267371,2.862261)(4.154078,2.862261)(4.081834,2.879821)(4.046645,2.879821)(4.012059,2.933818)(3.978058,2.970955)(3.847630,3.009044)(3.785571,3.009044)(3.667272,3.048122)(3.638844,3.068045)(3.529405,3.108681)(3.426357,3.215143)(3.259798,3.237316)(3.108681,3.305710)(3.088229,3.352935)(2.970955,3.426357)(2.827776,3.477118)(2.745093,3.529405)(2.496866,3.667272)(2.382796,3.978058)(2.058820,4.267371)(1.900449,4.557387)(1.719454,6.903101)(2.077039,inf)
        };
    \addlegendentry{ \unet }

    \addplot[,only marks,mark=*,green!75!black]
        coordinates {
            (3.951467,2.120741)
        };
    \addlegendentry{ SegEM~\cite{Berning2015} }

    \end{axis}
\end{tikzpicture}
    }
    \vspace{7mm}
    \caption{IED split/merge on \segem (higher is better).}
    \label{fig:results:split_merge:segem}
  \end{subfigure}
  \begin{subfigure}{0.5\textwidth}
    \vspace{1cm}
    \def\plotwidth{0.9\textwidth}
    \centerline{
      \subimport{../figures/results/plots/}{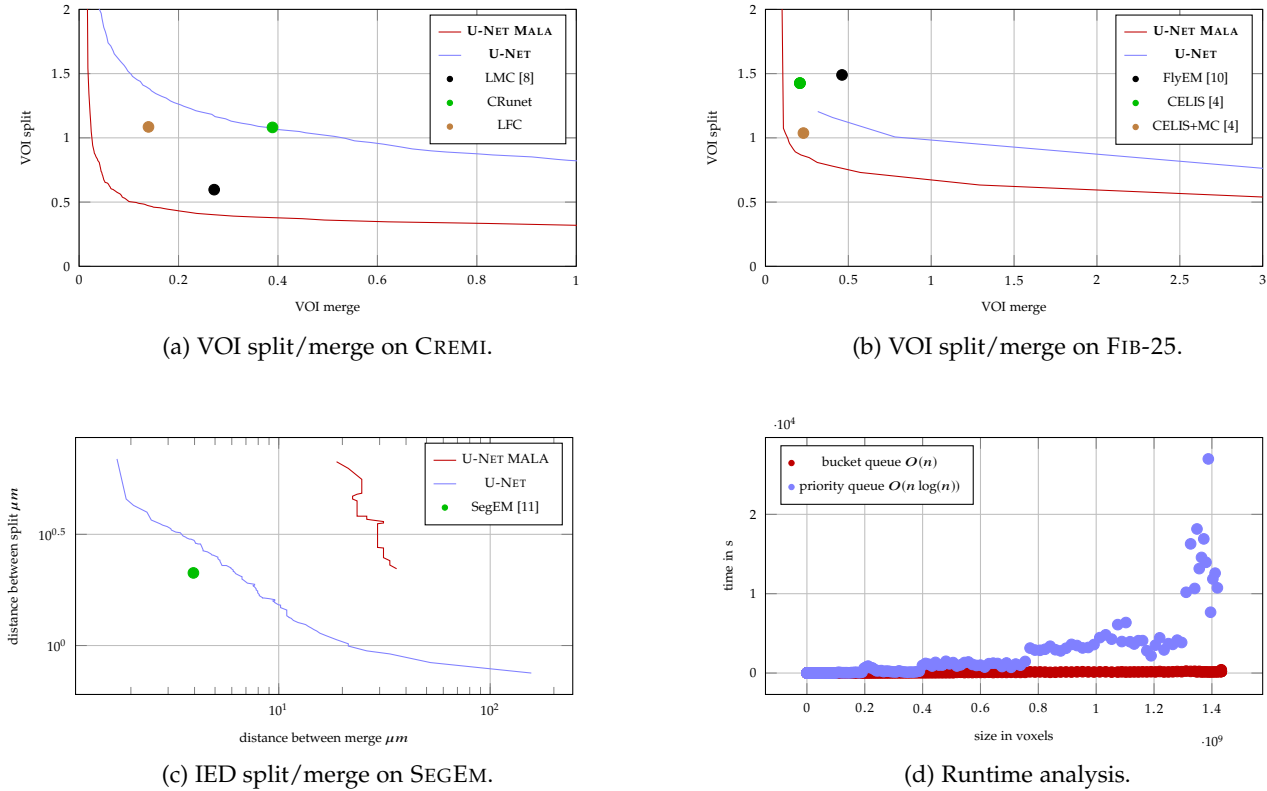}
    }
    \vspace{7mm}
    \caption{Runtime analysis.}
    \label{fig:results:performance}
  \end{subfigure}
  \caption{(a-c) Split merge curves of our method (lines) for different
  thresholds on the \cremi, \fib, and \segem datasets, compared against the
  best-ranking competing methods (dots). (d) Performance comparison of a naive
  agglomeration scheme (priority queue, $O(n\log(n))$) versus our linear-time
  agglomeration (bucket queue, $O(n)$).}
  \label{fig:results:split_merge}
\end{figure*}

% CREMI results detail figure
%
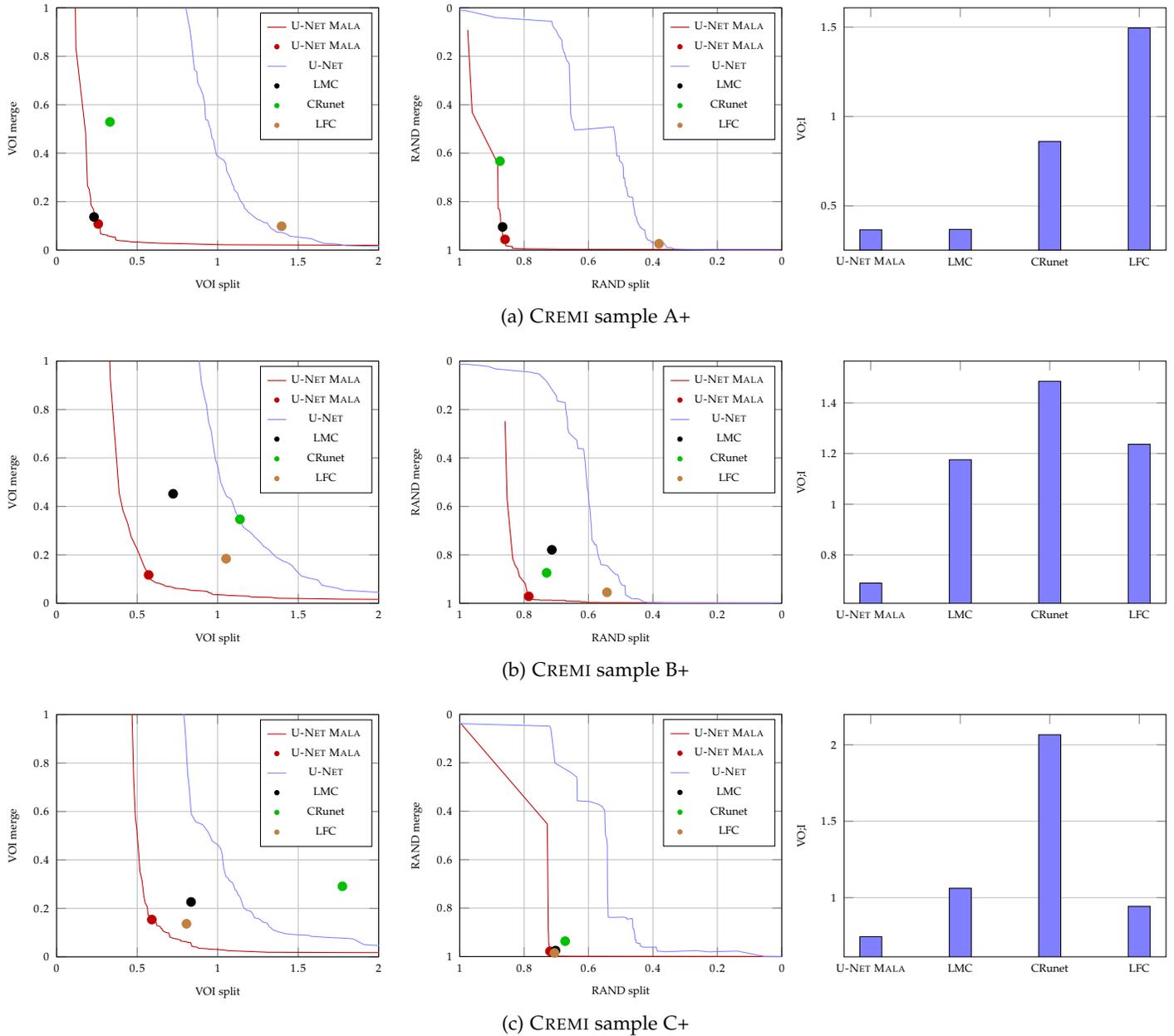
\begin{figure*}[t]
  \def\plotwidth{0.37\textwidth}
  \def\plotheight{0.3\textwidth}
  \begin{subfigure}{\textwidth}
    \centerline{
      \hspace{5mm}
      \begin{tikzpicture}
    \begin{axis}[
        ymajorgrids=true,
        xmajorgrids=true,
        width=\plotwidth,
        height=\plotheight,
        legend style={font=\tiny},
        legend image post style={scale=0.5},
        legend columns=1,
        xlabel=VOI split,
        ylabel=VOI merge,
        xticklabel style={font=\tiny,overlay},
        yticklabel style={font=\tiny,overlay},
        ylabel style={font=\tiny,overlay},
        xlabel style={font=\tiny,overlay},,xmin=0.000000,xmax=2.000000,ymin=0.000000,ymax=1.000000
    ]
        
    \addplot[,red!75!black]
        coordinates {
            (8.024109,0.006409)(1.043372,0.021807)(0.888475,0.024321)(0.775899,0.026648)(0.715954,0.027392)(0.660201,0.028071)(0.624613,0.028846)(0.595445,0.030181)(0.569372,0.030640)(0.558098,0.031020)(0.533227,0.031528)(0.530073,0.031814)(0.525103,0.032151)(0.520449,0.032545)(0.503919,0.032871)(0.497643,0.033031)(0.477683,0.033644)(0.462950,0.033892)(0.460102,0.034339)(0.455755,0.034880)(0.450698,0.035452)(0.447195,0.035910)(0.445838,0.036177)(0.437975,0.036268)(0.432864,0.036377)(0.426344,0.037507)(0.413352,0.038050)(0.400935,0.038228)(0.394872,0.038803)(0.389898,0.038822)(0.386355,0.039106)(0.384402,0.039310)(0.378365,0.039761)(0.377415,0.039992)(0.375609,0.040384)(0.373193,0.040767)(0.367511,0.042461)(0.365948,0.042699)(0.364925,0.053188)(0.356401,0.053365)(0.353252,0.053437)(0.348685,0.053802)(0.344287,0.054143)(0.343325,0.054903)(0.339005,0.055030)(0.335969,0.056610)(0.329120,0.057156)(0.328009,0.057423)(0.323899,0.057797)(0.323039,0.057983)(0.321399,0.058245)(0.317332,0.058471)(0.316578,0.058650)(0.314054,0.061671)(0.313380,0.061822)(0.311571,0.062261)(0.310482,0.062296)(0.303079,0.062561)(0.301723,0.062760)(0.298312,0.063460)(0.295726,0.063901)(0.290255,0.063987)(0.289194,0.064243)(0.288699,0.064265)(0.287919,0.064472)(0.287282,0.064618)(0.282082,0.065485)(0.281331,0.065791)(0.280576,0.065882)(0.278396,0.066167)(0.277569,0.066224)(0.275887,0.066746)(0.274999,0.067827)(0.273767,0.068209)(0.272148,0.068381)(0.270821,0.085594)(0.267913,0.095124)(0.266908,0.095416)(0.262971,0.096286)(0.262226,0.096433)(0.260156,0.099608)(0.258285,0.107769)(0.253688,0.109476)(0.253172,0.109923)(0.251892,0.120059)(0.249441,0.122597)(0.237007,0.136255)(0.235796,0.137209)(0.234607,0.152390)(0.233557,0.154202)(0.229160,0.167067)(0.222389,0.175621)(0.212947,0.186362)(0.210890,0.215423)(0.201573,0.252446)(0.191602,0.264623)(0.186688,0.359641)(0.180762,0.480101)(0.120037,0.833635)(0.086040,2.241645)
        };
    \addlegendentry{ \unet \mala }

    \addplot[,only marks,mark=*,red!75!black]
        coordinates {
            (0.258285,0.107769)
        };
    \addlegendentry{ \unet \mala }

    \addplot[,blue!50!white]
        coordinates {
            (8.186688,0.003086)(3.537682,0.005385)(2.806037,0.007374)(2.353110,0.009890)(2.192436,0.012190)(1.991366,0.014925)(1.912162,0.016365)(1.796314,0.018347)(1.741543,0.025087)(1.659942,0.027516)(1.615903,0.037605)(1.581377,0.047023)(1.529038,0.050941)(1.508476,0.052972)(1.449724,0.056912)(1.435330,0.063668)(1.408781,0.071408)(1.382961,0.073355)(1.366346,0.074573)(1.345490,0.084590)(1.331151,0.089598)(1.306227,0.112646)(1.288851,0.113511)(1.261311,0.124000)(1.248352,0.126397)(1.237162,0.131094)(1.219516,0.141156)(1.208285,0.147525)(1.188768,0.157144)(1.182657,0.162998)(1.166557,0.172851)(1.162646,0.186454)(1.147418,0.197434)(1.142237,0.202063)(1.129801,0.220083)(1.123195,0.234978)(1.103849,0.247637)(1.098909,0.263714)(1.092045,0.269483)(1.085130,0.285351)(1.081728,0.293017)(1.063289,0.317955)(1.055788,0.326387)(1.051690,0.355158)(1.037886,0.371276)(1.016451,0.378688)(0.997738,0.389557)(0.993105,0.390488)(0.979040,0.435165)(0.975945,0.451447)(0.963703,0.464027)(0.956524,0.498495)(0.949548,0.509933)(0.938761,0.536275)(0.925110,0.538550)(0.919886,0.601683)(0.909581,0.639154)(0.896037,0.663247)(0.875580,0.692487)(0.870853,0.735329)(0.856392,0.744846)(0.849374,0.782962)(0.840524,0.841954)(0.836732,0.864234)(0.828867,0.907701)(0.815342,0.952796)(0.799915,1.010683)(0.748358,1.066770)(0.738199,1.094419)(0.725395,1.156080)(0.719978,1.172993)(0.714595,1.197584)(0.701588,1.232986)(0.691935,1.293122)(0.688962,1.311660)(0.676144,1.465782)(0.662117,1.681419)(0.648273,1.722481)(0.635636,1.789597)(0.623315,1.849351)(0.597937,2.029700)(0.586235,2.066445)(0.579581,2.214338)(0.567321,2.289568)(0.556379,2.537748)(0.521185,2.693987)(0.513550,2.764829)(0.492686,3.158573)(0.464998,3.452441)(0.447801,3.802806)(0.321831,4.183268)(0.310331,4.447126)(0.180208,5.398873)(0.171125,5.603518)(0.080123,6.629734)(0.011458,7.396935)(0.004913,7.526022)(0.001954,7.658616)(0.000656,7.677981)(0.000000,7.681350)
        };
    \addlegendentry{ \unet }

    \addplot[,only marks,mark=*,black]
        coordinates {
            (0.232052,0.136558)
        };
    \addlegendentry{ LMC~\cite{Beier2017} }

    \addplot[,only marks,mark=*,green!75!black]
        coordinates {
            (0.331105,0.529078)
        };
    \addlegendentry{ CRunet~\cite{Zeng2017} }

    \addplot[,only marks,mark=*,brown]
        coordinates {
            (1.396811,0.098380)
        };
    \addlegendentry{ LFC~\cite{Parag2017} }

    \end{axis}
\end{tikzpicture}
      \hspace{10mm}
      \begin{tikzpicture}
    \begin{axis}[
        ymajorgrids=true,
        xmajorgrids=true,
        width=\plotwidth,
        height=\plotheight,
        legend style={font=\tiny},
        legend image post style={scale=0.5},
        legend columns=1,
        xlabel=RAND split,
        ylabel=RAND merge,
        xticklabel style={font=\tiny,overlay},
        yticklabel style={font=\tiny,overlay},
        ylabel style={font=\tiny,overlay},
        xlabel style={font=\tiny,overlay},,x dir=reverse,xmin=0.000000,xmax=1.000000,y dir=reverse,ymin=0.000000,ymax=1.000000
    ]
        
    \addplot[,red!75!black]
        coordinates {
            (0.003698,0.997501)(0.681320,0.996564)(0.716568,0.996021)(0.745942,0.995334)(0.759601,0.995245)(0.772473,0.995189)(0.788554,0.994946)(0.794147,0.994645)(0.799717,0.994590)(0.801541,0.994563)(0.806196,0.994512)(0.806745,0.994480)(0.807894,0.994437)(0.808334,0.994407)(0.811596,0.994337)(0.812531,0.994317)(0.816665,0.994237)(0.820040,0.994199)(0.820401,0.994144)(0.821039,0.993892)(0.821496,0.993824)(0.822111,0.993745)(0.822231,0.993704)(0.823425,0.993681)(0.824240,0.993663)(0.825316,0.993373)(0.827484,0.993329)(0.830482,0.993306)(0.831813,0.993225)(0.832516,0.993226)(0.832950,0.993183)(0.833134,0.993164)(0.834064,0.993031)(0.834169,0.993007)(0.834501,0.992927)(0.834811,0.992879)(0.836181,0.992651)(0.836369,0.992626)(0.836516,0.985170)(0.838349,0.985168)(0.838716,0.985166)(0.839984,0.984994)(0.840360,0.984955)(0.840495,0.984902)(0.841224,0.984893)(0.841749,0.984454)(0.842896,0.984373)(0.843099,0.984346)(0.847420,0.984301)(0.847566,0.984277)(0.847803,0.984226)(0.848413,0.984204)(0.848519,0.984150)(0.848838,0.983366)(0.848928,0.983350)(0.849179,0.983290)(0.849951,0.983284)(0.852682,0.983175)(0.852782,0.983157)(0.853181,0.982593)(0.853419,0.982505)(0.854780,0.982473)(0.854949,0.982457)(0.854976,0.982455)(0.855035,0.982433)(0.855066,0.982419)(0.856415,0.982274)(0.856540,0.982218)(0.856597,0.982195)(0.856797,0.982153)(0.856900,0.982146)(0.857046,0.982040)(0.857123,0.981954)(0.857211,0.981894)(0.857373,0.981856)(0.857562,0.967961)(0.858218,0.962099)(0.858414,0.961907)(0.858797,0.961364)(0.858872,0.961352)(0.859002,0.960716)(0.859126,0.956003)(0.860242,0.955432)(0.860326,0.955334)(0.860460,0.950912)(0.860850,0.950470)(0.866351,0.944510)(0.866530,0.943941)(0.866707,0.935457)(0.866817,0.934750)(0.867502,0.925897)(0.868865,0.921799)(0.871870,0.916601)(0.873387,0.860879)(0.876147,0.833487)(0.880626,0.828547)(0.881063,0.754292)(0.881919,0.642492)(0.961180,0.431729)(0.975013,0.091598)
        };
    \addlegendentry{ \unet \mala }

    \addplot[,only marks,mark=*,red!75!black]
        coordinates {
            (0.859126,0.956003)
        };
    \addlegendentry{ \unet \mala }

    \addplot[,blue!50!white]
        coordinates {
            (0.003371,0.998841)(0.132749,0.998902)(0.207201,0.998489)(0.251564,0.997810)(0.277153,0.997359)(0.306868,0.996738)(0.319738,0.996268)(0.335529,0.995374)(0.342498,0.990114)(0.356081,0.989829)(0.360233,0.982665)(0.365056,0.972632)(0.373817,0.968517)(0.375577,0.968277)(0.404364,0.969181)(0.405873,0.963607)(0.409619,0.959493)(0.412167,0.959316)(0.415468,0.959361)(0.419453,0.950231)(0.421165,0.949176)(0.425705,0.916801)(0.427818,0.916962)(0.432966,0.912116)(0.434689,0.911920)(0.436092,0.910296)(0.438173,0.907360)(0.439639,0.904415)(0.442576,0.897409)(0.443134,0.895868)(0.445804,0.892371)(0.446061,0.886533)(0.448715,0.881765)(0.449228,0.880672)(0.450484,0.869935)(0.451047,0.863988)(0.454784,0.858775)(0.455164,0.851866)(0.456222,0.850719)(0.457059,0.837042)(0.457361,0.834936)(0.460795,0.813737)(0.461482,0.810173)(0.461835,0.788314)(0.464138,0.780986)(0.473389,0.780769)(0.477358,0.775706)(0.477911,0.775753)(0.481412,0.745875)(0.481795,0.741201)(0.485354,0.738219)(0.486099,0.723877)(0.486895,0.719343)(0.487993,0.702408)(0.490985,0.703180)(0.491774,0.666247)(0.493688,0.646611)(0.495961,0.640634)(0.503946,0.630899)(0.504516,0.611575)(0.511861,0.611120)(0.513074,0.593038)(0.514504,0.566181)(0.515147,0.556514)(0.516928,0.540017)(0.518811,0.514471)(0.522434,0.491535)(0.642880,0.504713)(0.644490,0.497410)(0.647746,0.475273)(0.649701,0.472258)(0.650273,0.469281)(0.652601,0.458800)(0.654393,0.443518)(0.655071,0.439104)(0.657145,0.340061)(0.659560,0.233876)(0.662961,0.223363)(0.667051,0.219141)(0.669219,0.211334)(0.674328,0.185577)(0.676829,0.184280)(0.677981,0.172715)(0.680864,0.164690)(0.682612,0.133926)(0.690167,0.119609)(0.692796,0.115009)(0.699993,0.094292)(0.710240,0.077404)(0.713993,0.056052)(0.891849,0.039951)(0.895661,0.037488)(0.957280,0.018940)(0.959728,0.017964)(0.980741,0.012358)(0.998122,0.009080)(0.999576,0.008831)(0.999822,0.008586)(0.999942,0.008559)(1.000000,0.008555)
        };
    \addlegendentry{ \unet }

    \addplot[,only marks,mark=*,black]
        coordinates {
            (0.866755,0.904487)
        };
    \addlegendentry{ LMC~\cite{Beier2017} }

    \addplot[,only marks,mark=*,green!75!black]
        coordinates {
            (0.874884,0.633247)
        };
    \addlegendentry{ CRunet~\cite{Zeng2017} }

    \addplot[,only marks,mark=*,brown]
        coordinates {
            (0.381210,0.973913)
        };
    \addlegendentry{ LFC~\cite{Parag2017} }

    \end{axis}
\end{tikzpicture}
      \hspace{7mm}
      \begin{tikzpicture}
    \begin{axis}[
        symbolic x coords={\unet \mala,LMC~\cite{Beier2017},CRunet~\cite{Zeng2017},LFC~\cite{Parag2017}},
        xtick=data,
        ymajorgrids=true,
        width=\plotwidth,
        height=\plotheight,
        legend style={font=\tiny},
        legend image post style={scale=0.5},
        legend columns=2,
        ylabel=VO;I,,
        xticklabel style={font=\tiny,overlay},
        yticklabel style={font=\tiny,overlay},
        ylabel style={font=\tiny,overlay},
        xlabel style={font=\tiny,overlay},
    ]

        \addplot[ybar,fill=blue!50!white] coordinates {
            (\unet \mala,0.366053)(LMC~\cite{Beier2017},0.368610)(CRunet~\cite{Zeng2017},0.860183)(LFC~\cite{Parag2017},1.495191)
        };

    \end{axis}
\end{tikzpicture}
    }
    \vspace{7mm}
    \caption{\cremi sample A+}
  \end{subfigure}
  \begin{subfigure}{\textwidth}
    \vspace{5mm}
    \centerline{
      \hspace{5mm}
      \begin{tikzpicture}
    \begin{axis}[
        ymajorgrids=true,
        xmajorgrids=true,
        width=\plotwidth,
        height=\plotheight,
        legend style={font=\tiny},
        legend image post style={scale=0.5},
        legend columns=1,
        xlabel=VOI split,
        ylabel=VOI merge,
        xticklabel style={font=\tiny,overlay},
        yticklabel style={font=\tiny,overlay},
        ylabel style={font=\tiny,overlay},
        xlabel style={font=\tiny,overlay},,xmin=0.000000,xmax=2.000000,ymin=0.000000,ymax=1.000000
    ]
        
    \addplot[,red!75!black]
        coordinates {
            (7.774936,0.008038)(2.461866,0.014275)(2.122484,0.015748)(1.890976,0.016694)(1.772639,0.017031)(1.660913,0.018291)(1.612968,0.018783)(1.546974,0.019337)(1.508718,0.019817)(1.446982,0.020198)(1.404859,0.020435)(1.380122,0.020641)(1.352991,0.021813)(1.332496,0.024517)(1.306257,0.024666)(1.294240,0.024804)(1.265812,0.024975)(1.248873,0.025247)(1.238744,0.025425)(1.222764,0.025696)(1.211262,0.025793)(1.189276,0.029603)(1.181373,0.029798)(1.169566,0.030070)(1.144834,0.030442)(1.130812,0.030726)(1.119536,0.031158)(1.114463,0.031285)(1.099586,0.031676)(1.092906,0.031905)(1.077433,0.032269)(1.064409,0.032500)(1.054037,0.033067)(1.041315,0.034225)(1.022952,0.034718)(1.012506,0.035247)(1.006139,0.035437)(0.998369,0.035653)(0.987916,0.035965)(0.977525,0.036204)(0.972582,0.036421)(0.965464,0.037748)(0.956923,0.042578)(0.950885,0.043428)(0.941757,0.044047)(0.939063,0.048386)(0.932921,0.048988)(0.927796,0.049152)(0.916104,0.050258)(0.912986,0.050462)(0.902211,0.050840)(0.894366,0.051259)(0.890585,0.051445)(0.883036,0.052055)(0.855786,0.052626)(0.850479,0.053104)(0.846027,0.053341)(0.834210,0.053689)(0.827958,0.054221)(0.822125,0.055166)(0.817988,0.055506)(0.810459,0.057519)(0.802739,0.058121)(0.797672,0.058608)(0.793849,0.058980)(0.791132,0.059417)(0.785495,0.059963)(0.757487,0.060813)(0.743084,0.061309)(0.735138,0.061895)(0.729996,0.062939)(0.724211,0.065249)(0.719817,0.065638)(0.712465,0.067597)(0.706061,0.068647)(0.685430,0.069739)(0.678546,0.070955)(0.671435,0.071765)(0.663637,0.076684)(0.656590,0.077517)(0.647979,0.079697)(0.644658,0.080486)(0.629363,0.083645)(0.615941,0.084980)(0.602302,0.091168)(0.590664,0.094376)(0.580736,0.100770)(0.571299,0.117199)(0.562937,0.129532)(0.549929,0.142674)(0.542106,0.153856)(0.529420,0.175253)(0.515114,0.198694)(0.501048,0.221604)(0.462859,0.274186)(0.441152,0.329860)(0.411870,0.382683)(0.388020,0.456064)(0.332872,0.928030)(0.288568,1.976800)
        };
    \addlegendentry{ \unet \mala }

    \addplot[,only marks,mark=*,red!75!black]
        coordinates {
            (0.571299,0.117199)
        };
    \addlegendentry{ \unet \mala }

    \addplot[,blue!50!white]
        coordinates {
            (8.259425,0.004970)(4.878339,0.006307)(3.963595,0.006938)(3.258140,0.008192)(2.983378,0.008737)(2.680306,0.013848)(2.544864,0.015213)(2.370469,0.017233)(2.276630,0.018891)(2.149686,0.026985)(2.045252,0.039580)(1.997381,0.045387)(1.899014,0.049272)(1.855540,0.051033)(1.780786,0.053974)(1.742024,0.062825)(1.698353,0.067652)(1.647548,0.074394)(1.622292,0.096298)(1.584036,0.101543)(1.561051,0.104780)(1.521368,0.112156)(1.489130,0.133504)(1.460069,0.157000)(1.434272,0.167568)(1.401881,0.175629)(1.378696,0.185358)(1.364046,0.189765)(1.333132,0.210257)(1.313476,0.223921)(1.283676,0.236082)(1.269945,0.248495)(1.242678,0.265843)(1.227173,0.275993)(1.204579,0.292400)(1.161949,0.312376)(1.151396,0.324431)(1.137745,0.347420)(1.113682,0.373415)(1.086378,0.425977)(1.078822,0.432507)(1.054833,0.442364)(1.038438,0.467449)(1.018756,0.500544)(1.000013,0.568778)(0.986650,0.592053)(0.970666,0.660406)(0.960216,0.712231)(0.943256,0.750224)(0.931453,0.817645)(0.922139,0.840881)(0.908398,0.882444)(0.899463,0.910276)(0.876392,1.057863)(0.871000,1.084812)(0.857446,1.195682)(0.841936,1.235306)(0.816506,1.270700)(0.797411,1.370853)(0.782364,1.406391)(0.749855,1.523603)(0.737413,1.570465)(0.730340,1.620217)(0.717188,1.678002)(0.707003,1.868390)(0.701416,1.906582)(0.687633,2.023556)(0.678512,2.143395)(0.673093,2.166099)(0.660455,2.284774)(0.656782,2.324045)(0.645861,2.515679)(0.622715,2.621029)(0.614186,2.722204)(0.611296,2.758237)(0.605285,2.887393)(0.532447,3.398398)(0.523613,3.504776)(0.463911,3.890599)(0.407459,4.084891)(0.253293,4.474801)(0.185001,4.978328)(0.176405,5.038300)(0.137802,5.256323)(0.108720,5.616242)(0.090609,5.835825)(0.072433,5.946135)(0.065525,6.060107)(0.059439,6.200724)(0.035287,6.650662)(0.010425,6.815418)(0.003971,7.041724)(0.003508,7.109517)(0.000910,7.276218)(0.000685,7.343845)(0.000497,7.393492)(0.000162,7.577012)(0.000001,7.629525)(0.000000,7.647946)(0.000000,7.650113)
        };
    \addlegendentry{ \unet }

    \addplot[,only marks,mark=*,black]
        coordinates {
            (0.723330,0.452069)
        };
    \addlegendentry{ LMC~\cite{Beier2017} }

    \addplot[,only marks,mark=*,green!75!black]
        coordinates {
            (1.138565,0.346739)
        };
    \addlegendentry{ CRunet~\cite{Zeng2017} }

    \addplot[,only marks,mark=*,brown]
        coordinates {
            (1.052431,0.184003)
        };
    \addlegendentry{ LFC~\cite{Parag2017} }

    \end{axis}
\end{tikzpicture}
      \hspace{10mm}
      \begin{tikzpicture}
    \begin{axis}[
        ymajorgrids=true,
        xmajorgrids=true,
        width=\plotwidth,
        height=\plotheight,
        legend style={font=\tiny},
        legend image post style={scale=0.5},
        legend columns=1,
        xlabel=RAND split,
        ylabel=RAND merge,
        xticklabel style={font=\tiny,overlay},
        yticklabel style={font=\tiny,overlay},
        ylabel style={font=\tiny,overlay},
        xlabel style={font=\tiny,overlay},,x dir=reverse,xmin=0.000000,xmax=1.000000,y dir=reverse,ymin=0.000000,ymax=1.000000
    ]
        
    \addplot[,red!75!black]
        coordinates {
            (0.005539,0.997975)(0.451832,0.997430)(0.490185,0.997330)(0.513968,0.997073)(0.551479,0.997202)(0.569516,0.996759)(0.573713,0.996730)(0.580683,0.996689)(0.585762,0.996659)(0.594489,0.996658)(0.601050,0.996663)(0.603141,0.996639)(0.606113,0.995416)(0.608440,0.993725)(0.611308,0.993727)(0.612617,0.993709)(0.617147,0.993706)(0.620013,0.993656)(0.620720,0.993648)(0.622665,0.993630)(0.623543,0.993635)(0.626509,0.992148)(0.626898,0.992126)(0.628557,0.992113)(0.639045,0.992036)(0.640982,0.992047)(0.641899,0.992027)(0.642323,0.992003)(0.643638,0.991978)(0.643978,0.991955)(0.646794,0.991795)(0.650438,0.991791)(0.651699,0.991708)(0.656891,0.991549)(0.661732,0.991551)(0.662799,0.991518)(0.663169,0.991518)(0.663830,0.991491)(0.665874,0.991467)(0.666789,0.991451)(0.667356,0.991372)(0.668788,0.991198)(0.669897,0.989633)(0.670424,0.989568)(0.671632,0.989437)(0.671805,0.988156)(0.672266,0.988135)(0.672532,0.988118)(0.673857,0.987661)(0.674028,0.987642)(0.675707,0.987619)(0.677512,0.987610)(0.677829,0.987606)(0.678971,0.987479)(0.704538,0.987741)(0.705188,0.987727)(0.706257,0.987723)(0.712091,0.987787)(0.712815,0.987715)(0.713275,0.987605)(0.713526,0.987594)(0.714445,0.987117)(0.715294,0.987077)(0.715678,0.987041)(0.716059,0.986987)(0.716284,0.986960)(0.717480,0.986905)(0.739962,0.986933)(0.748514,0.986818)(0.751203,0.986823)(0.751663,0.986674)(0.752117,0.986349)(0.752512,0.986281)(0.753264,0.986020)(0.753842,0.985636)(0.766061,0.985694)(0.766997,0.985557)(0.767804,0.985500)(0.768704,0.984347)(0.770308,0.984306)(0.771842,0.983963)(0.772150,0.983918)(0.775274,0.983610)(0.779012,0.983421)(0.781160,0.982410)(0.782922,0.982142)(0.784572,0.980106)(0.785514,0.971423)(0.786860,0.966766)(0.788275,0.959352)(0.789245,0.956681)(0.790799,0.950271)(0.792918,0.940168)(0.796620,0.917786)(0.814231,0.889639)(0.820253,0.857814)(0.828055,0.842341)(0.835263,0.815104)(0.852730,0.568818)(0.858976,0.248046)
        };
    \addlegendentry{ \unet \mala }

    \addplot[,only marks,mark=*,red!75!black]
        coordinates {
            (0.785514,0.971423)
        };
    \addlegendentry{ \unet \mala }

    \addplot[,blue!50!white]
        coordinates {
            (0.003939,0.998624)(0.189796,0.998728)(0.271959,0.998843)(0.334352,0.998509)(0.354601,0.998476)(0.389164,0.997373)(0.397653,0.997177)(0.410704,0.996806)(0.417434,0.996545)(0.429131,0.993323)(0.441610,0.983931)(0.444948,0.982027)(0.457510,0.981030)(0.460395,0.980838)(0.467902,0.980265)(0.472697,0.973313)(0.476994,0.971927)(0.485224,0.965486)(0.486878,0.927190)(0.490242,0.926717)(0.491953,0.925660)(0.497195,0.924051)(0.503814,0.906427)(0.506995,0.889059)(0.509599,0.881563)(0.516247,0.880673)(0.517986,0.877652)(0.519033,0.877178)(0.527203,0.866896)(0.528890,0.862847)(0.532701,0.859501)(0.534026,0.856034)(0.537531,0.851335)(0.539187,0.847778)(0.541712,0.844537)(0.561917,0.839875)(0.562576,0.837513)(0.563512,0.826053)(0.569426,0.806241)(0.575832,0.758825)(0.576265,0.757551)(0.581443,0.757786)(0.583198,0.748278)(0.589314,0.739030)(0.591380,0.672320)(0.592408,0.666797)(0.594550,0.623497)(0.595857,0.605691)(0.598401,0.587340)(0.600505,0.543602)(0.601103,0.540461)(0.602732,0.530150)(0.603712,0.524398)(0.610732,0.410749)(0.611240,0.407146)(0.614342,0.369516)(0.616201,0.362871)(0.632456,0.360980)(0.636345,0.325973)(0.639984,0.322661)(0.657157,0.301255)(0.661729,0.294190)(0.662252,0.287535)(0.664455,0.278765)(0.665372,0.237205)(0.665869,0.234522)(0.668342,0.218679)(0.668793,0.210121)(0.669302,0.209077)(0.670801,0.197796)(0.671016,0.195192)(0.672465,0.171469)(0.697269,0.164513)(0.698078,0.157788)(0.698226,0.157141)(0.699069,0.145423)(0.729512,0.085114)(0.731574,0.081129)(0.753545,0.053425)(0.780709,0.045647)(0.889632,0.032142)(0.911551,0.021838)(0.913005,0.021668)(0.925221,0.020184)(0.938132,0.017748)(0.949914,0.016934)(0.953744,0.016495)(0.955937,0.016136)(0.957705,0.015633)(0.974530,0.013503)(0.998286,0.013169)(0.999472,0.012384)(0.999584,0.012235)(0.999923,0.011744)(0.999955,0.011635)(0.999960,0.011520)(0.999985,0.010959)(1.000000,0.010846)(1.000000,0.010814)(1.000000,0.010811)
        };
    \addlegendentry{ \unet }

    \addplot[,only marks,mark=*,black]
        coordinates {
            (0.713703,0.779243)
        };
    \addlegendentry{ LMC~\cite{Beier2017} }

    \addplot[,only marks,mark=*,green!75!black]
        coordinates {
            (0.729945,0.874159)
        };
    \addlegendentry{ CRunet~\cite{Zeng2017} }

    \addplot[,only marks,mark=*,brown]
        coordinates {
            (0.542680,0.954706)
        };
    \addlegendentry{ LFC~\cite{Parag2017} }

    \end{axis}
\end{tikzpicture}
      \hspace{7mm}
      \begin{tikzpicture}
    \begin{axis}[
        symbolic x coords={\unet \mala,LMC~\cite{Beier2017},CRunet~\cite{Zeng2017},LFC~\cite{Parag2017}},
        xtick=data,
        ymajorgrids=true,
        width=\plotwidth,
        height=\plotheight,
        legend style={font=\tiny},
        legend image post style={scale=0.5},
        legend columns=2,
        ylabel=VO;I,,
        xticklabel style={font=\tiny,overlay},
        yticklabel style={font=\tiny,overlay},
        ylabel style={font=\tiny,overlay},
        xlabel style={font=\tiny,overlay},
    ]

        \addplot[ybar,fill=blue!50!white] coordinates {
            (\unet \mala,0.688497)(LMC~\cite{Beier2017},1.175399)(CRunet~\cite{Zeng2017},1.485304)(LFC~\cite{Parag2017},1.236434)
        };

    \end{axis}
\end{tikzpicture}
    }
    \vspace{7mm}
    \caption{\cremi sample B+}
  \end{subfigure}
  \begin{subfigure}{\textwidth}
    \vspace{5mm}
    \centerline{
      \hspace{5mm}
      \begin{tikzpicture}
    \begin{axis}[
        ymajorgrids=true,
        xmajorgrids=true,
        width=\plotwidth,
        height=\plotheight,
        legend style={font=\tiny},
        legend image post style={scale=0.5},
        legend columns=1,
        xlabel=VOI split,
        ylabel=VOI merge,
        xticklabel style={font=\tiny,overlay},
        yticklabel style={font=\tiny,overlay},
        ylabel style={font=\tiny,overlay},
        xlabel style={font=\tiny,overlay},,xmin=0.000000,xmax=2.000000,ymin=0.000000,ymax=1.000000
    ]
        
    \addplot[,red!75!black]
        coordinates {
            (8.304157,0.002300)(1.313699,0.019115)(1.180379,0.022602)(1.066438,0.025751)(1.028636,0.028982)(0.979220,0.031305)(0.947652,0.032172)(0.929919,0.033970)(0.909791,0.034341)(0.882843,0.035116)(0.869179,0.040796)(0.859830,0.041380)(0.846856,0.041792)(0.840247,0.043567)(0.832387,0.058650)(0.828213,0.059004)(0.814742,0.059854)(0.809932,0.060079)(0.807475,0.061695)(0.798356,0.062984)(0.792378,0.063485)(0.780891,0.064332)(0.777014,0.064555)(0.770154,0.065699)(0.767851,0.070456)(0.760082,0.071088)(0.757025,0.071611)(0.751743,0.072262)(0.746977,0.074016)(0.743833,0.074134)(0.737537,0.074536)(0.735211,0.074976)(0.731151,0.075332)(0.727379,0.076559)(0.723688,0.076786)(0.716361,0.078492)(0.714315,0.078694)(0.702694,0.081863)(0.699936,0.092306)(0.696422,0.098403)(0.691504,0.098763)(0.688829,0.099206)(0.684210,0.099416)(0.682527,0.100456)(0.678751,0.100809)(0.677277,0.100904)(0.674925,0.101995)(0.673708,0.102410)(0.671817,0.103100)(0.668710,0.103546)(0.662749,0.105626)(0.660247,0.107661)(0.659186,0.110039)(0.657578,0.111452)(0.656398,0.111717)(0.654102,0.115458)(0.648246,0.117628)(0.645464,0.122846)(0.643238,0.123872)(0.640590,0.125235)(0.635942,0.126082)(0.635169,0.126274)(0.632892,0.127759)(0.631111,0.128120)(0.622986,0.129250)(0.620753,0.131773)(0.618226,0.141823)(0.615996,0.143024)(0.611832,0.143902)(0.606804,0.144977)(0.604876,0.145522)(0.600942,0.146204)(0.599540,0.147143)(0.594973,0.148501)(0.593605,0.152438)(0.591035,0.153814)(0.587881,0.155820)(0.586027,0.156708)(0.583649,0.157539)(0.580866,0.164568)(0.578118,0.167493)(0.575969,0.168545)(0.574063,0.170881)(0.567183,0.173474)(0.563683,0.204232)(0.556660,0.213943)(0.555292,0.215725)(0.552121,0.219093)(0.549001,0.223956)(0.540622,0.254915)(0.536652,0.278281)(0.530641,0.314005)(0.516414,0.355184)(0.510310,0.419048)(0.497304,0.530317)(0.488772,0.576853)(0.478435,0.732904)(0.469586,0.966345)(0.428186,2.118440)(0.034166,6.433399)
        };
    \addlegendentry{ \unet \mala }

    \addplot[,only marks,mark=*,red!75!black]
        coordinates {
            (0.591035,0.153814)
        };
    \addlegendentry{ \unet \mala }

    \addplot[,blue!50!white]
        coordinates {
            (9.115814,0.001160)(5.251753,0.005146)(4.256341,0.023378)(3.412528,0.028412)(3.036018,0.033453)(2.704182,0.037530)(2.503934,0.038488)(2.310737,0.040329)(2.202517,0.041828)(2.061506,0.045661)(1.965332,0.047843)(1.908823,0.050562)(1.841749,0.073944)(1.800815,0.076471)(1.714988,0.079165)(1.675343,0.080365)(1.634360,0.082412)(1.581695,0.084451)(1.563019,0.089035)(1.527373,0.090217)(1.501760,0.090756)(1.469203,0.092312)(1.448874,0.094056)(1.418305,0.095647)(1.403153,0.099270)(1.379685,0.102438)(1.358048,0.108541)(1.347647,0.109785)(1.319336,0.125887)(1.307041,0.139506)(1.281688,0.142746)(1.272561,0.148476)(1.256056,0.157438)(1.244919,0.158474)(1.229659,0.160896)(1.215888,0.179368)(1.206356,0.188005)(1.184059,0.192177)(1.178541,0.193198)(1.163290,0.201252)(1.155862,0.212757)(1.145265,0.237256)(1.131841,0.254593)(1.126004,0.256170)(1.118277,0.271852)(1.109616,0.274389)(1.094651,0.303731)(1.087292,0.308269)(1.071366,0.315416)(1.063784,0.325268)(1.050472,0.332126)(1.041989,0.356806)(1.038598,0.362632)(1.030837,0.407480)(1.027109,0.428424)(1.017969,0.447532)(1.001375,0.462367)(0.965028,0.478769)(0.962406,0.484898)(0.940342,0.515931)(0.906939,0.545849)(0.864277,0.557372)(0.837046,0.588446)(0.830982,0.632326)(0.824904,0.691645)(0.813609,0.757144)(0.799843,0.922144)(0.793533,0.975503)(0.785253,1.028465)(0.778838,1.109397)(0.770172,1.162283)(0.765795,1.200769)(0.758893,1.230857)(0.749719,1.313286)(0.745600,1.353977)(0.733029,1.529977)(0.714193,1.648139)(0.710301,1.697869)(0.686802,1.848294)(0.651463,1.951910)(0.596328,2.131687)(0.588183,2.390003)(0.583672,2.529015)(0.577100,2.614146)(0.573388,2.739879)(0.545297,2.977597)(0.536351,3.039208)(0.468019,3.486785)(0.438980,3.890392)(0.356181,5.050692)(0.323898,5.445435)(0.033894,6.228147)(0.017217,6.387675)(0.014960,6.587241)(0.011794,6.678944)(0.008411,6.716142)(0.002622,6.773766)(0.001317,6.791380)(0.000812,6.801324)(0.000000,6.809160)
        };
    \addlegendentry{ \unet }

    \addplot[,only marks,mark=*,black]
        coordinates {
            (0.834799,0.226486)
        };
    \addlegendentry{ LMC~\cite{Beier2017} }

    \addplot[,only marks,mark=*,green!75!black]
        coordinates {
            (1.774478,0.291116)
        };
    \addlegendentry{ CRunet~\cite{Zeng2017} }

    \addplot[,only marks,mark=*,brown]
        coordinates {
            (0.806378,0.136358)
        };
    \addlegendentry{ LFC~\cite{Parag2017} }

    \end{axis}
\end{tikzpicture}
      \hspace{10mm}
      \begin{tikzpicture}
    \begin{axis}[
        ymajorgrids=true,
        xmajorgrids=true,
        width=\plotwidth,
        height=\plotheight,
        legend style={font=\tiny},
        legend image post style={scale=0.5},
        legend columns=1,
        xlabel=RAND split,
        ylabel=RAND merge,
        xticklabel style={font=\tiny,overlay},
        yticklabel style={font=\tiny,overlay},
        ylabel style={font=\tiny,overlay},
        xlabel style={font=\tiny,overlay},,x dir=reverse,xmin=0.000000,xmax=1.000000,y dir=reverse,ymin=0.000000,ymax=1.000000
    ]
        
    \addplot[,red!75!black]
        coordinates {
            (0.002285,0.999560)(0.658868,0.998706)(0.667623,0.998432)(0.685643,0.998297)(0.689783,0.998130)(0.692841,0.997974)(0.695735,0.997938)(0.696705,0.997887)(0.697675,0.997867)(0.699615,0.997840)(0.700385,0.997133)(0.700909,0.997097)(0.701390,0.997071)(0.702297,0.996909)(0.702601,0.991980)(0.702737,0.991976)(0.703433,0.991947)(0.703594,0.991944)(0.703722,0.991898)(0.704306,0.991859)(0.704611,0.991844)(0.706084,0.991819)(0.706238,0.991817)(0.706551,0.991756)(0.706674,0.991149)(0.707166,0.991113)(0.707338,0.991102)(0.707660,0.991032)(0.708015,0.990869)(0.708261,0.990853)(0.708776,0.990844)(0.708846,0.990820)(0.709170,0.990800)(0.709449,0.990764)(0.709715,0.990760)(0.712297,0.989612)(0.712367,0.989593)(0.713411,0.989326)(0.713595,0.986893)(0.713873,0.984220)(0.714093,0.984198)(0.714195,0.984183)(0.714532,0.984178)(0.714574,0.984123)(0.714770,0.984053)(0.714812,0.984046)(0.714903,0.983734)(0.715008,0.983680)(0.715128,0.983616)(0.715278,0.983600)(0.715845,0.983291)(0.716039,0.982591)(0.716075,0.982359)(0.716190,0.982276)(0.716266,0.982262)(0.716344,0.982069)(0.716623,0.981897)(0.716713,0.981285)(0.716869,0.981264)(0.716940,0.981231)(0.717080,0.981182)(0.717109,0.981175)(0.717222,0.981091)(0.717304,0.981079)(0.717952,0.981031)(0.718007,0.980888)(0.718109,0.978930)(0.718239,0.978614)(0.718530,0.978524)(0.718696,0.978477)(0.718745,0.978452)(0.718972,0.978441)(0.719051,0.978419)(0.719231,0.978378)(0.719276,0.978216)(0.719417,0.978156)(0.719483,0.978021)(0.719702,0.978015)(0.719859,0.977976)(0.720001,0.977198)(0.720205,0.976904)(0.720300,0.976849)(0.720455,0.976762)(0.720985,0.976682)(0.721184,0.969731)(0.721461,0.968595)(0.721578,0.968535)(0.721730,0.968105)(0.721901,0.967723)(0.722192,0.958600)(0.722377,0.953377)(0.722643,0.947964)(0.723616,0.941564)(0.723858,0.922298)(0.724372,0.897100)(0.724782,0.889610)(0.725089,0.834099)(0.725385,0.783066)(0.727908,0.452590)(0.997738,0.035860)
        };
    \addlegendentry{ \unet \mala }

    \addplot[,only marks,mark=*,red!75!black]
        coordinates {
            (0.719417,0.978156)
        };
    \addlegendentry{ \unet \mala }

    \addplot[,blue!50!white]
        coordinates {
            (0.001219,0.999721)(0.055027,0.999071)(0.137321,0.977165)(0.228057,0.980371)(0.264329,0.975329)(0.291528,0.976307)(0.334046,0.978901)(0.353910,0.979773)(0.361173,0.980065)(0.375240,0.977909)(0.381190,0.977937)(0.386236,0.977922)(0.390200,0.960828)(0.394849,0.960757)(0.410096,0.961105)(0.412928,0.961137)(0.415556,0.961157)(0.422562,0.961577)(0.423588,0.960967)(0.426304,0.960994)(0.427895,0.961081)(0.429513,0.961032)(0.431108,0.960056)(0.433087,0.960071)(0.434030,0.958451)(0.435421,0.958245)(0.437592,0.957346)(0.438223,0.957273)(0.439856,0.950098)(0.440724,0.945669)(0.449108,0.944301)(0.449477,0.943476)(0.450343,0.942727)(0.451003,0.942642)(0.451910,0.942468)(0.452488,0.932491)(0.452994,0.931593)(0.454426,0.931294)(0.454748,0.931217)(0.455539,0.927295)(0.456063,0.924588)(0.456626,0.906569)(0.457300,0.901458)(0.457612,0.901247)(0.457965,0.896876)(0.458493,0.895377)(0.459394,0.886041)(0.459750,0.885702)(0.460791,0.885315)(0.461120,0.884349)(0.461804,0.883481)(0.462449,0.878782)(0.462618,0.878333)(0.462920,0.857817)(0.463077,0.853600)(0.463581,0.847391)(0.464951,0.842983)(0.480644,0.845612)(0.480792,0.845200)(0.482260,0.839514)(0.492471,0.837097)(0.525604,0.838745)(0.539034,0.837382)(0.539359,0.822095)(0.539739,0.786672)(0.540778,0.752888)(0.541734,0.540955)(0.542329,0.530130)(0.542898,0.524950)(0.543650,0.511110)(0.546246,0.503229)(0.546647,0.500245)(0.547420,0.497782)(0.547967,0.486207)(0.548091,0.481441)(0.549739,0.396587)(0.556204,0.384182)(0.556530,0.380459)(0.570530,0.370790)(0.595933,0.359976)(0.634869,0.357079)(0.635492,0.275862)(0.635613,0.267928)(0.636026,0.263755)(0.636154,0.259216)(0.649141,0.246099)(0.650176,0.243976)(0.704416,0.200980)(0.708046,0.153195)(0.716992,0.057485)(0.721165,0.048683)(0.997466,0.038662)(0.998710,0.036788)(0.998772,0.034694)(0.998991,0.034083)(0.999169,0.033870)(0.999790,0.033556)(0.999828,0.033459)(0.999920,0.033413)(1.000000,0.033374)
        };
    \addlegendentry{ \unet }

    \addplot[,only marks,mark=*,black]
        coordinates {
            (0.702403,0.975194)
        };
    \addlegendentry{ LMC~\cite{Beier2017} }

    \addplot[,only marks,mark=*,green!75!black]
        coordinates {
            (0.672341,0.936446)
        };
    \addlegendentry{ CRunet~\cite{Zeng2017} }

    \addplot[,only marks,mark=*,brown]
        coordinates {
            (0.705579,0.984859)
        };
    \addlegendentry{ LFC~\cite{Parag2017} }

    \end{axis}
\end{tikzpicture}
      \hspace{7mm}
      \begin{tikzpicture}
    \begin{axis}[
        symbolic x coords={\unet \mala,LMC~\cite{Beier2017},CRunet~\cite{Zeng2017},LFC~\cite{Parag2017}},
        xtick=data,
        ymajorgrids=true,
        width=\plotwidth,
        height=\plotheight,
        legend style={font=\tiny},
        legend image post style={scale=0.5},
        legend columns=2,
        ylabel=VO;I,,
        xticklabel style={font=\tiny,overlay},
        yticklabel style={font=\tiny,overlay},
        ylabel style={font=\tiny,overlay},
        xlabel style={font=\tiny,overlay},
    ]

        \addplot[ybar,fill=blue!50!white] coordinates {
            (\unet \mala,0.744850)(LMC~\cite{Beier2017},1.061285)(CRunet~\cite{Zeng2017},2.065594)(LFC~\cite{Parag2017},0.942735)
        };

    \end{axis}
\end{tikzpicture}
    }
    \vspace{7mm}
    \caption{\cremi sample C+}
  \end{subfigure}
  \caption{Comparison of the proposed method against competing methods on the
  \cremi testing datasets A+, B+, and C+. Shown are (from left to right)
  variation of information (VOI, split and merge contribution), Rand index
  (RAND, split and merge contribution), and the total VOI (split and merge
  combined). Baseline \emph{\unet} is our method, but without \malis
  training (\ie, only minimizing the Euclidean distance to the ground-truth
  affinities during training) For \unet \mala, the red dot indicates the best
  threshold found on the training data.
  }
  \label{fig:supplemental:results:cremi}
\end{figure*}

\begin{table*}[t]
  \rowcolors{2}{gray!2!white}{gray!20!white}
  \centerline{
  \begin{tabular}{l|ccc >{\raggedright}p{3.5cm} >{\raggedright\arraybackslash}p{3cm}}
    Name & Imaging & Tissue & Resolution & Training Data & Testing Data \\
    \hline
    \cremi
      & ssTEM
      & \emph{Drosophila}
      & $4\x4\x40$\,nm
      & 3 volumes of $1250\x1250\x125$ voxels
      & 3 volumes of $1250\x1250\x125$ voxels \\
    \fib
      & FIBSEM
      & \emph{Drosophila}
      & $8\x8\x8$\,nm
      & $520\x520\x520$ voxels
      & $13.8$ gigavoxels \\
    \segem
      & SBEM 	% SegEM is NOT FIBSEM
      & mouse cortex
      & $11\x11\x26$\,nm
      & 279 volumes of $100\x100\x100$ voxels
      & $400\x400\x350$ voxels (skeletons) \\
  \end{tabular}
  }
  \caption{Overview of used datasets.}
  \label{tab:results:datasets}
\end{table*}

\noindent{\bf Datasets}
We present results on three different and diverse datasets:
  \cremi~\cite{Cremi}, \fib~\cite{Takemura2015}, and
  \segem~\cite{Berning2015} (see \tabref{tab:results:datasets} for an
  overview).
  These datasets sum up to almost 15 gigavoxels of testing data, with \fib
  alone contributing 13.8 gigavoxels, thus challenging automatic segmentation
  methods for their efficiency. In fact, only two methods have so far been
  evaluated on \fib~\cite{Takemura2015,Maitin-Shepard2016}.
  Another challenge is posed by the \cremi dataset: Coming from serial section
  EM, this dataset is highly anisotropic and contains artifacts like support
  film folds, missing sections, and staining precipitations.
  %
  % i added 1 pixel to the receptive fields -- you forgot the center pixel! :-)
  Regardless of the differences in isotropy and presence of artifacts, we use
  the same method (3D \unet training, prediction, and agglomeration) for all
  datasets. The size of the receptive field of the \unet was set for each
  dataset to be approximately one \textmu m$^3$, \ie, $213\x213\x29$ for \cremi,
  $89\x89\x89$ for \fib, and $89\x89\x49$ for \segem.
  For the \cremi dataset, we also pre-aligned training and testing data with an
  elastic alignment method~\cite{Saalfeld2012}, using the padded volumes
  provided by the challenge.

\noindent{\bf Training}
We implemented and trained our network using the \caffe library on modestly
augmented training data for which we performed random rotations, transpositions
and flips, as well as elastic deformations.
  On the anisotropic \cremi dataset, we further simulated missing sections by
  setting intensity values to 0 ($p=0.05$) and low contrast sections by
  multiplying the intensity variance by $0.5$ ($p=0.05$).
  We used the Adam optimizer~\cite{Kingma2014} with an initial learning rate of
  $\alpha=10^{-4}$, $\beta_1=0.95$, $\beta_2=0.99$, and $\epsilon=10^{-8}$.

\noindent{\bf Quantitative results}
On each of the investigated datasets, we see a clear improvement in
accuracy using our method, compared to the current state of the art.
  We provide quantitative results for each of the datasets individually, where
  we compare our method (labeled \unet \mala) against different other
  methods\footnote{\revised The presented results reflect the state of the
  \cremi challenge at the time of writing, see~\cite{Cremi}.}. We also include
  a baseline (labeled \unet) in our analysis, which is our method, but trained
  without the \constmalis loss.
  In \tabref{tab:results}, we report the segmentation obtained on the best
  threshold found in the respective training datasets. In
  \figref{fig:results:split_merge}, we show the split/merge curve for varying
  thresholds of our agglomeration scheme.

For \segem, we do not use the metric proposed by Berning et
al.~\cite{Berning2015}, as we found it to be problematic:
  The authors suggest an overlap threshold of 2 to compensate for inaccuracies
  in the ground-truth, however this has the unintended consequence of ignoring
  some neurons in the ground-truth for poor segmentations.
  For the \segem segmentation (kindly provided by the authors), 195 out of 225
  ground-truth skeletons are ignored because of insufficient overlap with any
  segmentation label. On our segmentation, only 70 skeletons would be ignored,
  thus the results are not directly comparable.
  Therefore, we performed a new IED evaluation using TED~\cite{Funke2017}, a
  metric that allows slight displacement of skeleton nodes (we chose 52\,nm in
  this case) in an attempt to minimize splits and merges. This metric reveals
  that our segmentations (\unet \mala) improve over both split and merge
  errors, over all thresholds of agglomeration, including the initial fragments
  (see~\figref{fig:results:split_merge:segem}).

\noindent{\bf Qualitative results}
\begin{figure*}
  \centerline{\includegraphics[width=0.85\textwidth]{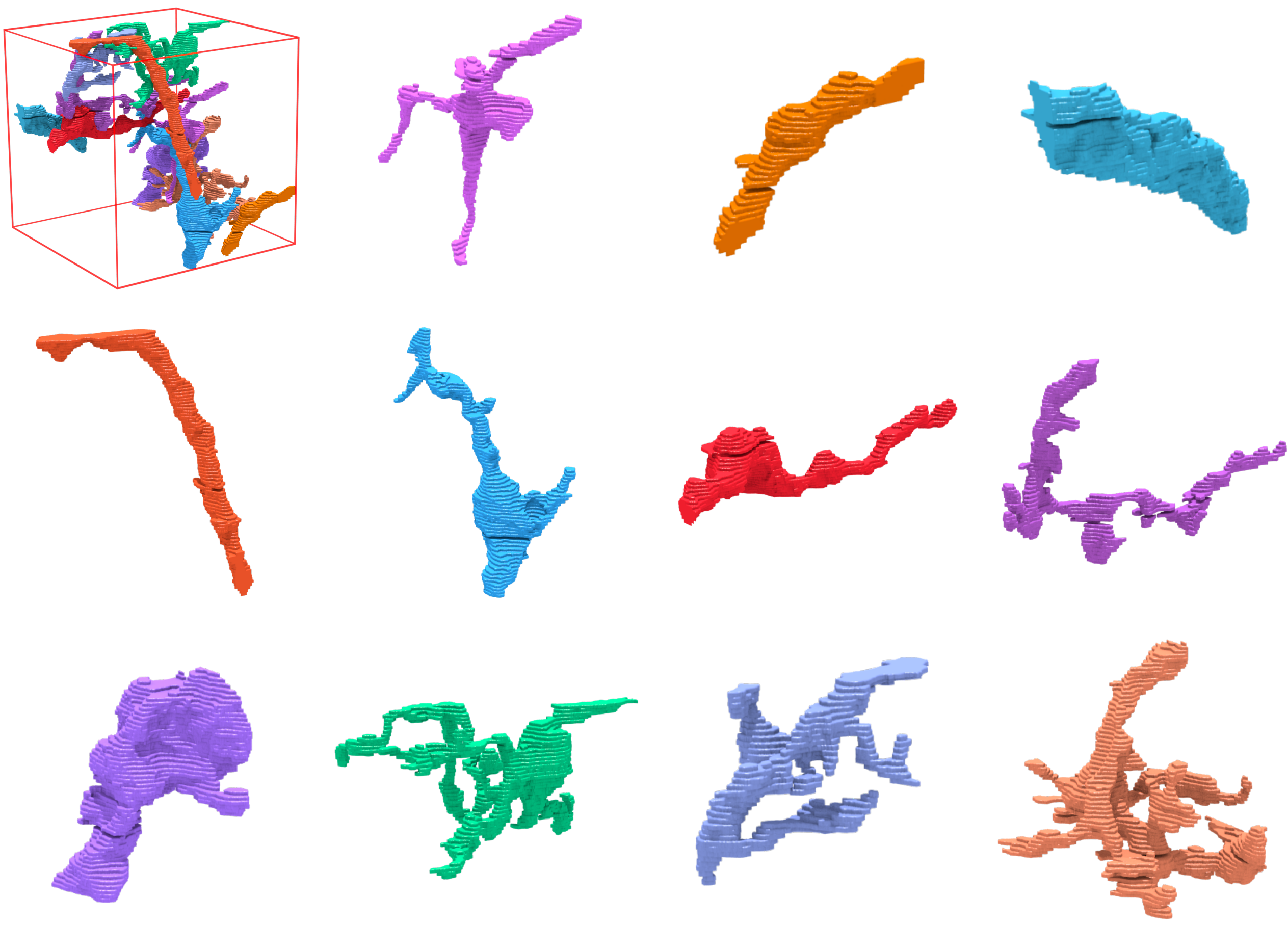}}
  \caption{Reconstructions of 11 randomly selected neurons of the 100 largest
  found in the \cremi test volume C+.}
  \label{fig:results:qual:cremi}
\end{figure*}
\begin{figure*}
  \centerline{\includegraphics[width=0.85\textwidth]{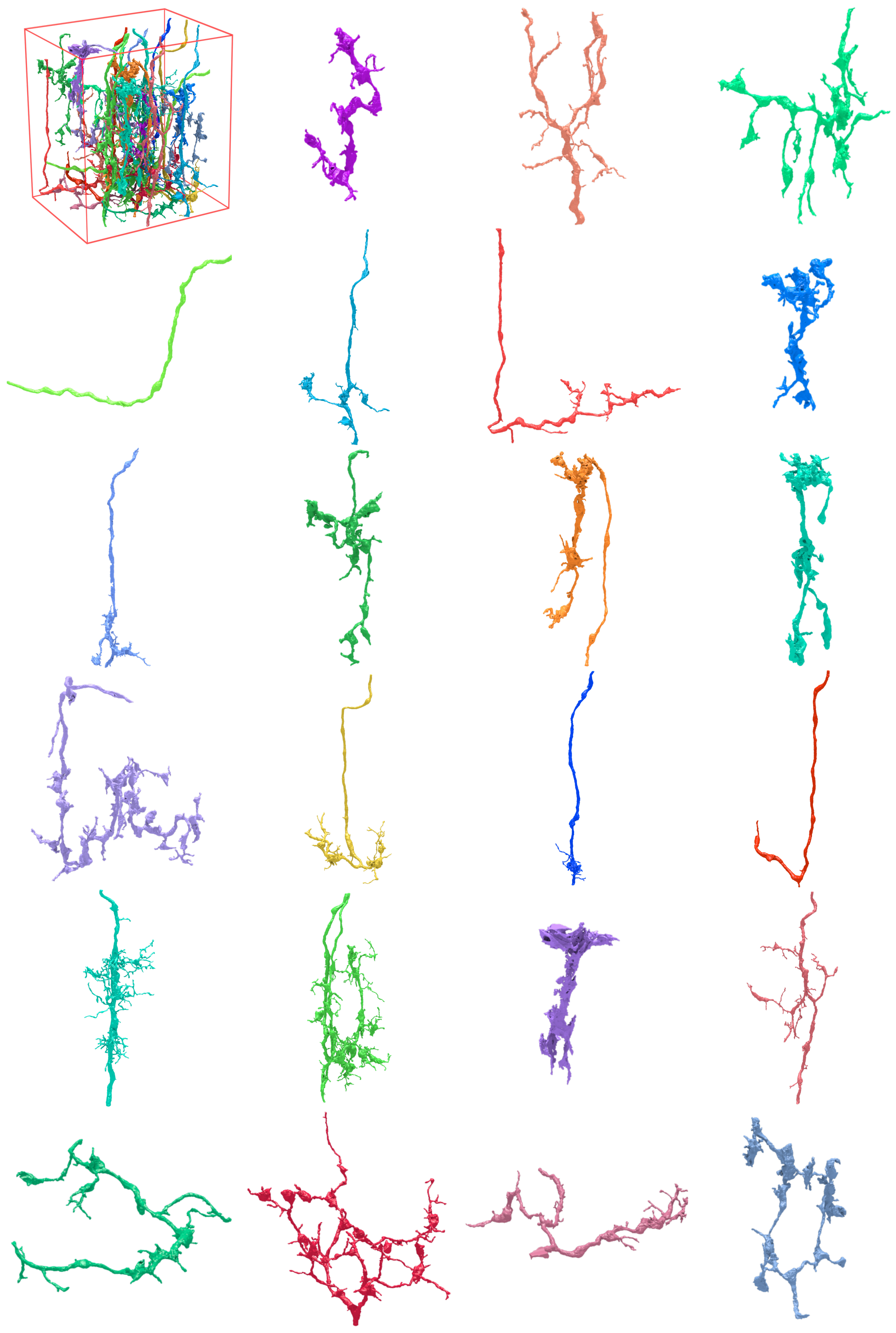}}
  \caption{Reconstructions of 23 randomly selected neurons of the 500 largest
  found in the \fib test volume.}
  \label{fig:results:qual:fib}
\end{figure*}
Renderings of 11 and 23 randomly selected neurons, reconstructed using the
proposed method, are shown for the test regions of \cremi and \fib in
\figref{fig:results:qual:cremi} and \figref{fig:results:qual:fib},
respectively.

\noindent{\bf Dataset (an)isotropy}
Save for minor changes in the network architectures and the generation of
initial fragments, our method works unchanged on both near-isotropic block-face
datasets (\fib, \segem) as well as on highly anisotropic serial-section
datasets (\cremi).
  These findings suggest that there is no need for specialized constructions
  like dedicated features for anisotropic volumes or separate classifiers
  trained for merging of fragments within or across sections.

\noindent{\bf Merge functions}
\label{sec:results:mergefunction}
\begin{table}
  \begin{subtable}{0.5\textwidth}
    \centerline{\rowcolors{2}{gray!2!white}{gray!20!white}
\begin{tabular}{l|dddd}
    &\multicolumn{1}{c}{VOI split}&\multicolumn{1}{c}{VOI merge}&\multicolumn{1}{c}{VOI sum}&\multicolumn{1}{c}{CREMI score}\\

    \hline
15\%&0.583&0.063&0.646&0.212\\
25\%&0.441&0.092&0.533&0.188\\
50\%&0.397&0.056&0.453&0.156\\
75\%&0.347&0.074&\multicolumn{1}{B{.}{.}{-1} }{0.421}&\multicolumn{1}{B{.}{.}{-1} }{0.146}\\
85\%&0.347&0.084&0.431&0.156\\
mean&0.380&0.058&0.438&0.149\\
Zlateski~\cite{Zlateski2015}&1.015&1.010&2.025&0.364
\end{tabular}
}
    \caption{\cremi (training data).}
    \label{tab:results:merge_function:cremi}
  \end{subtable}
  \begin{subtable}{0.5\textwidth}
    \vspace{7mm}
    \centerline{\rowcolors{2}{gray!2!white}{gray!20!white}
\begin{tabular}{l|D{.}{.}{3}D{.}{.}{3}D{.}{.}{3}}
    &\multicolumn{1}{c}{VOI split}&\multicolumn{1}{c}{VOI merge}&\multicolumn{1}{c}{VOI sum}\\

    \hline
15\%&1.480&0.364&1.844\\
25\%&1.393&0.163&1.555\\
50\%&1.115&0.234&\multicolumn{1}{B{.}{.}{3} }{1.350}\\
75\%&1.085&0.318&1.402\\
85\%&1.176&0.394&1.570\\
mean&1.221&0.198&1.418\\
Zlateski~\cite{Zlateski2015}&1.054&1.017&2.071
\end{tabular}
}
    \caption{\fib.}
    \label{tab:results:merge_function:cremi}
  \end{subtable}
  \vspace{3mm}
  \caption{Results for different merge functions of our method compared with
  the agglomeration strategy proposed in~\cite{Zlateski2015}. We show the
  results at the threshold achieving the best score in the respective dataset
  (CREMI score for \cremi, VOI for \fib). Note that, for this analysis, we used
  the available training datasets which explains deviations from the numbers
  shown in~\tabref{tab:results}.}
  \label{tab:results:merge_function}
\end{table}
Our method for efficient agglomeration allows using a range of different merge
functions.
  In \tabref{tab:results:merge_function}, we show results for different choices
  of quantile merge functions, mean affinity, and an agglomeration baseline
  proposed in~\cite{Zlateski2015} on datasets \cremi and \fib. Even across
  these very different datasets, we see best results for affinity quantiles
  between 50\% and 75\%. {\rerevised All initial edge scores have been set to
  one minus the maximum predicted affinity between the regions. Theoretically,
  this is the ideal choice since the \malis training optimizes the maximin
  affinity between regions. Also empirically we found this initialization to
  perform consistently better than others (like the mean or a quantile
  affinity).}

\noindent{\bf Throughput}
\tabref{tab:results:throughput} shows the throughput of our method for each
dataset, broken down into affinity prediction (\unet), fragment extraction
(watershed), and fragment agglomeration (agglomeration).
  For \cremi and \segem, most time is spent on the prediction of affinities.
  The faster predictions in \fib are due to less feature maps used in the
  network for this dataset.

To empirically confirm the theoretical speedup of using a bucket queue for
agglomeration, we show in \figref{fig:results:performance} a speed comparison
of the proposed linear-time agglomeration against a naive agglomeration scheme
for volumes of different sizes.

\begin{table}[t]
  \rowcolors{2}{gray!2!white}{gray!20!white}
  \centerline{
    {
    \begin{tabular}{l|cccc}
      dataset & \unet & watershed & agglomeration & total \\
      \hline
      \cremi
        & $3.04$
        & $0.23$
        & $0.83$
        & $4.10$ \\
      \fib
        & $0.66$
        & $0.92$
        & $1.28$
        & $2.86$ \\
      \segem
        & $2.19$
        & $0.25$
        & $0.14$
        & $2.58$ \\
    \end{tabular}
    }
  }
  \caption{Throughput of our method for each of the investigated datasets in
  seconds per megavoxel.}
  \label{tab:results:throughput}
\end{table}

\section{Discussion}

%% Oh, I hate it when discussions start like that. So I won't. Let's cut to the meat.
%%  -Jan
%
%We presented a method for neuron segmentation on 3D EM datasets, which both
%improves in speed and accuracy compared to previous methods.

A remarkable property of the \mala method is that it requires almost no tuning
to operate on datasets of different characteristics, except for minor changes
in the size of the receptive field of the \unet, training data augmentation to
model dataset specific artifacts, and initial fragment generation.
  This suggests that there is no need for the development of dedicated
  algorithms for different EM modalities. Across all datasets, our results indicate that affinity
  predictions on voxels are sufficiently accurate to render sophisticated post-precessing obsolete.
  It remains an open question whether fundamentally different approaches, like
  the recently reported flood-filling network~\cite{Januszewski2016}, also
  generalize in a similar way. At the time of writing, neither code nor data
  were publicly available for a direct comparison.

Furthermore, the \unet is the only part in our method that requires training,
so that all training data can (and should) be used to correctly predict
affinities.
  This is an advantage over current state-of-the-art methods that require careful
  splitting of precious training data into non-overlapping sets used to train
  voxel-wise predictions and an agglomeration classifier (or
  accepting the disadvantages of having the sets overlap).

Although linear in runtime and memory, correct parallelization of hierarchical
agglomeration is not trivial and will require further research.
  However, as demonstrated on the FIB-25 dataset, naive block-based
  agglomeration followed by empirical stitching based on region overlap
  generates very satisfying practical results.

%\subimport{sections/}{acknowledgements}

{
\bibliographystyle{naturemag}
\bibliography{references}
}

%\newpage

%\appendix

%\subimport{sections/}{supplemental}

\end{document}